# A hierarchical deep learning framework for the consistent classification of land use objects in geospatial databases

C. Yang *, F. Rottensteiner, C. Heipke

Institute of Photogrammetry and GeoInformation, Leibniz Universität Hannover - Germany
{yang, rottensteiner, heipke}@ipi.uni-hannover.de

Abstract:

Land use as contained in geospatial databases constitutes an essential input for different applications such as urban management, regional planning and environmental monitoring. In this paper, a hierarchical deep learning framework is proposed to verify the land use information. For this purpose, a two-step strategy is applied. First, given high-resolution aerial images, the land cover information is determined. To achieve this, an encoder-decoder based convolutional neural network (CNN) is proposed. Second, the pixel-wise land cover information along with the aerial images serves as input for another CNN to classify land use. Because the object catalogue of geospatial databases is frequently constructed in a hierarchical manner, we propose a new CNN-based method aiming to predict land use in multiple levels *hierarchically* and *simultaneously*. A so called *Joint Optimization (JO)* is proposed where predictions are made by selecting the hierarchical tuple over all levels which has the maximum joint class scores, providing *consistent* results across the different levels.

The conducted experiments show that the CNN relying on *JO* outperforms previous results, achieving an overall accuracy up to 92.5%. In addition to the individual experiments on two test sites, we investigate whether data showing different characteristics can improve the results of land cover and land use classification, when processed together. To do so, we combine the two datasets and undertake some additional experiments. The results show that adding more data helps both land cover and land use classification, especially the identification of underrepresented categories, despite their different characteristics.

# 1. Introduction

Information on *land use*, i.e. on the social and economic function of a parcel, is typically stored in geospatial databases, where each parcel is represented as a polygonal object and has a label indicating its land use class. Land use provides high benefit for many applications, for instance in urban management, regional planning, and environmental monitoring. Because of fast changes of land use due to e.g. urban growth and agricultural land use conversion, these geospatial databases become outdated quickly. To keep them up-to-date, the content can be compared to new remote sensing data. If there is a contradiction between the new data and the database content for a specific object, the object in the database needs to be updated. The database verification process requires an enormous effort, so that its automation is desirable.

Database verification is typically implemented as a two-step process: first, land cover per pixel is determined; afterwards, the land cover information along with the aerial images and possibly additional information such as height data is combined in a second classification process to determine the land use for every database object (e.g. Gerke et al., 2008; Helmholz et al., 2012; Hermosilla et al., 2012; Montanges et al., 2015; Albert et al., 2017; Zhang et al., 2018; Zhang et al., 2019). This strategy is chosen because land use objects of a given class (e.g. *residential, agricultural*) are often composed of different land cover elements, so that the prediction of land cover delivers valuable information for determining land use. Classification is normally applied in a supervised manner, most recently mainly based on Convolutional Neural Networks (CNN; Krizhevsky et al., 2012), e.g. (Zhang et al., 2018; Yang et al., 2019).

Many existing methods for land use classification dedicated to verifying geospatial databases differentiate only a relatively small number of rather broadly defined classes. For instance, Albert et al. (2017) distinguish ten land use classes covering urban and rural areas; Zhang et al. (2018) focus solely on urban areas while also classifying ten classes. Yang and Newsam (2010) differentiate 21 classes, but the definition of these classes is relatively broad and comparable to the papers just mentioned; the number of classes is larger mainly due to the fact that their test data cover a very large geographical region. However, the object catalogues of land use databases are much more fine-grained. For instance, in the land use layer of the German cadastre, about 190 categories are differentiated (AdV, 2008). Obviously, these different classes cannot all be separated using remote sensing imagery. Many geospatial databases contain land use information in different semantic levels of abstraction. At the coarsest level, there are only a few broad classes such as *settlement, traffic, vegetation* or *water body* to be differentiated. As the semantic resolution of categories increases, these classes are hierarchically refined from level to level; two examples are shown in Fig. 1.

In this paper, we propose a strategy for investigating and potentially improving the maximum semantic resolution that can be achieved in land use classification for database verification, based on image and height data with a ground sampling distance (GSD) in the order of a few decimetres. Our method follows the two-stage procedure of land cover and land use classification mentioned earlier, finally producing a set of land use labels for every object of an existing land use database, where each label corresponds to a different semantic level of the object catalogue. We propose to predict the land use categories of multiple semantic levels *simultaneously* using an approach based on CNN, considering the hierarchical structure of object catalogues. In this context, we extend our previous work (Yang et al., 2020a, 2020b) to achieve a hierarchical land use classification process that enforces *consistency* with the hierarchical object class catalogue. Our methods are tested in two sites located in Germany; the land cover classification network is also tested using the ISPRS labelling challenge (Wegner et al., 2017).

| Binary object mask | RGB Orthophoto | Size | Level | Category |
|---|---|---|---|---|
| 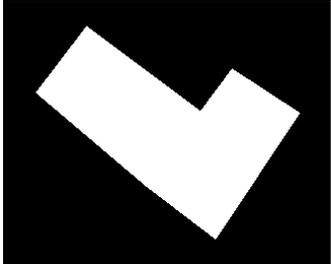 | 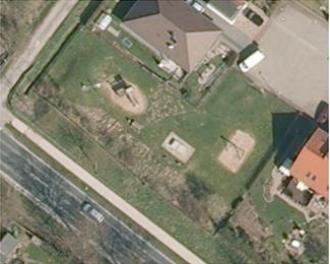 | 320 x 260 pixels (area in database: 980 m$^2$) | I | *settlement* |
| | | | II | *recreation* |
| | | | III | *leisure* |
| 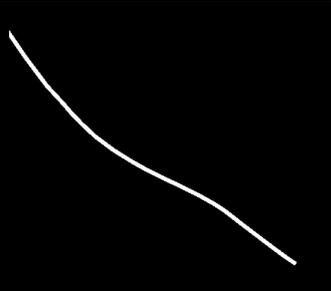 | 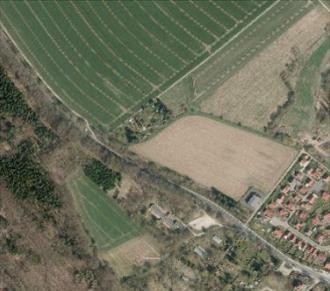 | 3400 x 3100 pixels (area in database: 6600 m$^2$) | I | *traffic* |
| | | | II | *road traffic* |
| | | | III | *motor road* |

Figure 1: Two database objects with corresponding RGB orthophotos (GSD: 20 cm) and categories in three semantic layers of (AdV, 2008), starting from the coarsest (I) to the finest one (III). The two objects are displayed at different scales. The object shape is represented by a binary object mask at the scale of the orthophoto.

The scientific contributions of this paper are the following:

- Our main contribution is a new strategy to achieve *hierarchical training and inference* of land use for objects from a geospatial database. We achieve this by predicting a tuple of class labels that is guaranteed to be consistent with the class hierarchy of the object catalogue and which has the maximum joint classification probability over all semantic levels of the catalogue. To train the CNN, a new loss function is proposed that respects the hierarchy of the class labels according to the object catalogue for each training object. To the best of our knowledge, this is the first method for considering this hierarchy in training and inference for land use classification based on CNN.

- We investigate potential overfitting of the classification results by employing CNNs of different complexity. We show experimentally that the number of parameters can be reduced considerably without significantly compromising the achieved accuracy.

- We conduct different experiments to demonstrate the advantages and limitations of our new approach, including a section which illustrates the advantages of a guaranteed hierarchy for the land use object labels.

- Inspired by Kaiser et al. (2017), we also compare the performance of land cover and land use clas-

sification for different definitions of the training and test sets. While the majority of our experiments is based on training and test sets from the same test site, we also combine the data of the test sites to investigate whether adding more training data helps the classification even if there is a domain gap (Tuia et al., 2016), i.e. if the distributions of the features and the classes of the two datasets differ from each other.

Section 2 contains a review of related work. Our deep learning framework is presented in section 3, where the first part gives an overview about the land cover classification and the second part describes our hierarchical land use classification, which is the main focus of this paper. Section 4 describes the test data and the test setup for the experimental evaluation of our approach, while the evaluation is presented in section 5. Conclusions and an outlook are given in section 6.

## 2. Related work

### 2.1. Land cover classification

For land cover classification a number of CNN-based approaches have been proposed in recent years. Fully Connected Networks (FCN, Long et al., 2015) or encoder-decoder based networks (Noh et al., 2015) are able to predict a separate prediction per pixel. *SegNet* (Badrinarayanan et al., 2017) and *U-Net* (Ronneberger et al., 2015) are also based on encoder-decoder networks and apply end-to-end learning of all parameters. These papers solve the problem of pixel-wise segmentation of images, a task called "semantic segmentation" in computer vision. This is the task to be solved in land cover classification, but none of these methods has originally been developed for that purpose; the classes that were differentiated in the original tests were unrelated to land cover. However, many papers have adapted such FCN for land cover classification and related tasks; cf. (Zhu et al., 2017; Ma et al., 2019) for reviews of applications of CNN in remote sensing. To mitigate the problem of such approaches with inaccurate boundaries inherent in these approaches, skip-connections are introduced e.g. in Zhao et al. (2017) and Lin et al. (2017). Promising results have been reported based on different variants of such networks, see e.g. Maggiori et al. (2017), Xie et al. (2017), Audebert et al. (2018) and Marmanis et al. (2018). In our previous work (Yang et al., 2019) we employed the *SegNet* architecture (Badrinarayanan et al., 2017) and introduced skip connections, which can be learnt (Maggiori et al., 2017).

### 2.2. Land use classification

In the context of classifying land use objects, a particular challenge for applying CNN is the large variation of object shape and size. To the best of our knowledge, the first work classifying land use objects from a geospatial database by a CNN is (Yang et al., 2018), where large polygons are decomposed into multiple patches that can be classified by a CNN. However, the method only considered image and land cover information inside the boundary polygon of each object, disregarding the outside grey values, which leads to a loss of context information. In a subsequent publication the authors considered the polygon as a mask to represent the polygon shape and size, but kept the complete image information (Yang et al., 2019). Gujrathi et al. (2020) integrated DenseNet (Huang et al. 2017) modules on the basis of (Yang et al., 2019) for one-level land use classification. Yang et al. (2020b) adapted the basic network architecture of Yang et al. (2019) so that it could consider class labels at multiple semantic levels, yet in a greedy way.

Zhang et al. (2018) proposed a method to classify urban land use objects based on CNN. They start with pre-processing the image using a (non-semantic) mean-shift segmentation (Comaniciu & Meer, 2002). Subsequently, the segmentation results are used to obtain polygons based on which the input for a CNN is generated. However, they focus solely on urban scenes, without any consideration of rural areas. Later, the authors suggested a strategy based on multi-layer perceptrons (MLP) for land cover and a CNN for land use to simultaneously classify land cover and land use in one step (Zhang et al., 2019). The authors show that they can separate 10 different land use classes.

### 2.3. Classification with hierarchical class structures

A CNN-based method dedicated to multi-label classification of aerial images is proposed by Hua et al. (2019). In their work, the authors predict multiple labels for an image, describing each object type appearing in the image. However, an explicit model of semantic relations between these labels is not used. As a consequence, the results are not guaranteed to be consistent with the class hierarchy. The relations between the different levels can also be taken into account in multi-task learning, see Leiva-Murillo et al. (2013) for an example.

Turning to computer vision, several approaches to solve hierarchical classification of images considering multiple semantic levels haven been proposed. The first CNN-based work for that purpose is presented by Deng et al. (2014). The authors introduce a so called Hierarchy and Exclusion (HEX) graph in which different types of semantic relations are modelled. Their proposed CNN takes the HEX graph as input for training and inference, outputs all class probabilities in one layer, and yields a hierarchically consistent result. However, the suggested training and inference procedure is very complex and time consuming. Other researchers, e.g. Guo et al. (2018), utilize a recurrent neural network (RNN) to address hierarchical classification while they consider the classification from coarse to fine level as sequential prediction, the other direction (fine to coarse) is not considered.

Bidirectional information passing of class scores is suggested in the work of Hu et al. (2016). They enhance the level-specific class scores by propagating information from other semantic levels. Nonetheless, they only consider the message passing between neighbouring levels. Though Hu et al. (2016) is embedded in a completely different application context, it inspired our work (Yang et al., 2020b), in which we propose *hierarchical training and inference* of land use. The main methodological extensions of this paper are a joint optimization procedure that enforces consistency of the predictions at different semantic levels with the hierarchy of the object catalogue of the geospatial database and a new training procedure that also considers this class hierarchy in an explicit way.

### 3. Methodology

In this paper, we present a new two-step strategy for hierarchical land use classification based on image and height data with the aim to verify geospatial databases. Fig. 2 shows an overview of the automatic verification process and visualises the principles of the hierarchical class structure. The input of our method consists of high-resolution multispectral aerial images, a DSM, a DTM and the geospatial database. The ground sampling distance (GSD) should be sufficient to differentiate the ob-ject classes required for the land cover and land use classification steps (typically, in the range of a few decimetres). The database is supposed to consist of objects represented by their boundary polygons and information about the land use. The object catalogue is supposed to be of a hierarchical nature and the land use of each polygon is supposed to be known at all semantic levels that are meaningful for a specific object category.

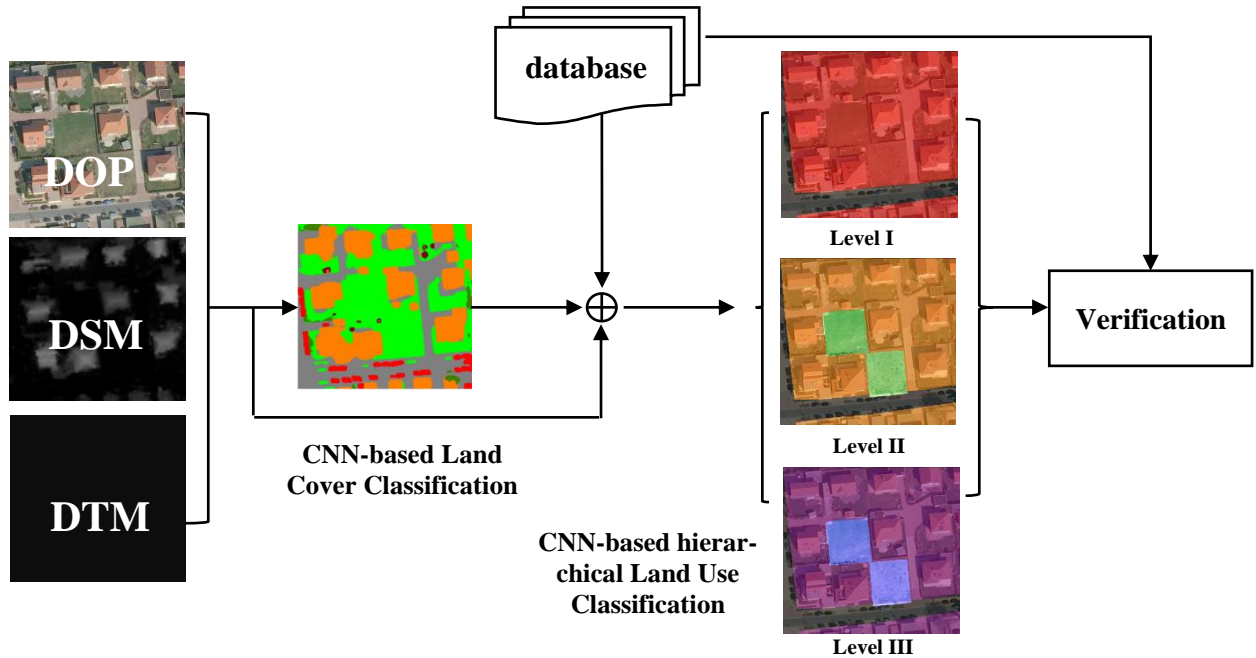

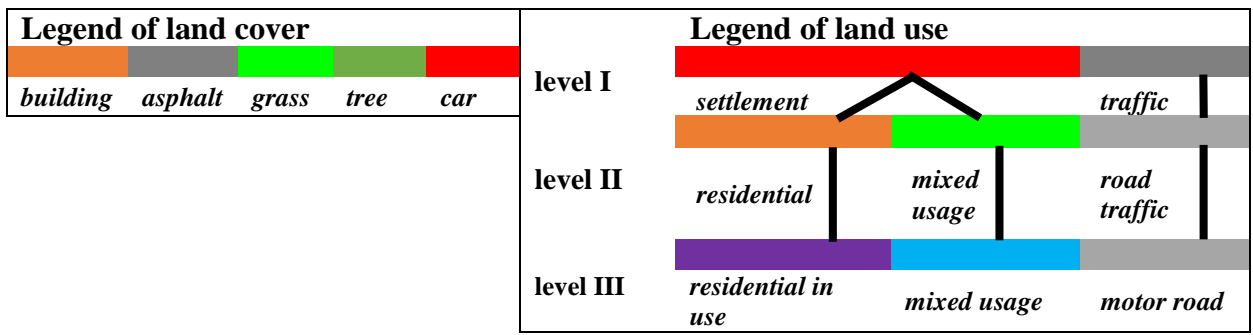

Figure 2: Overview of the hierarchical deep learning framework. The bold lines between the class labels at different semantic levels indicate the hierarchical relationships between the classes.

In the first step, described in Section 3.1, the image and height data are used to determine the land cover of every pixel by a CNN. Apart from the predicted land cover category, this results in a set of probabilistic class scores (one per land cover class) for every pixel. In the second step, which is the main methodological contribution of this paper and which is described in Section 3.2, these land cover class scores are used along with the image and height data to predict land use labels at multiple semantic levels of the object catalogue for each land use object in the database, again using a CNN-based technique. Each label corresponds to one semantic level of the object catalogue, and the predictions must be consistent with the class hierarchy, i.e. labels predicted for finer semantic levels have to be children of the categories predicted for the coarser levels of the hierarchy.

In the final verification step the predicted class labels can be compared to the land use information in the database: if the predicted labels are consistent with the database information, the object is considered to be unchanged, otherwise it is flagged as a potential error in the database that has to be inspected by a human operator. As this paper focusses on the two classification processes, the verification itself is not investigated further.

### 3.1. Land cover classification

### 3.1.1. Basic architecture and training procedure

For the pixel-wise classification of land cover we adapt the best-performing network of our previous work (Yang et al., 2020a), referred to as *FuseNet-234* in that publication, and for reasons of simplicity as *FuseNet* in this paper. In a variant of *FuseNet* we use the same basic architecture, but reduce the number of filters in each layer to produce a network having fewer parameters. This reduced network will be referred to as *FuseNet-lite* in the remainder of this paper. Fig. 3 shows the architecture of both, *FuseNet* and *FuseNet-lite*.

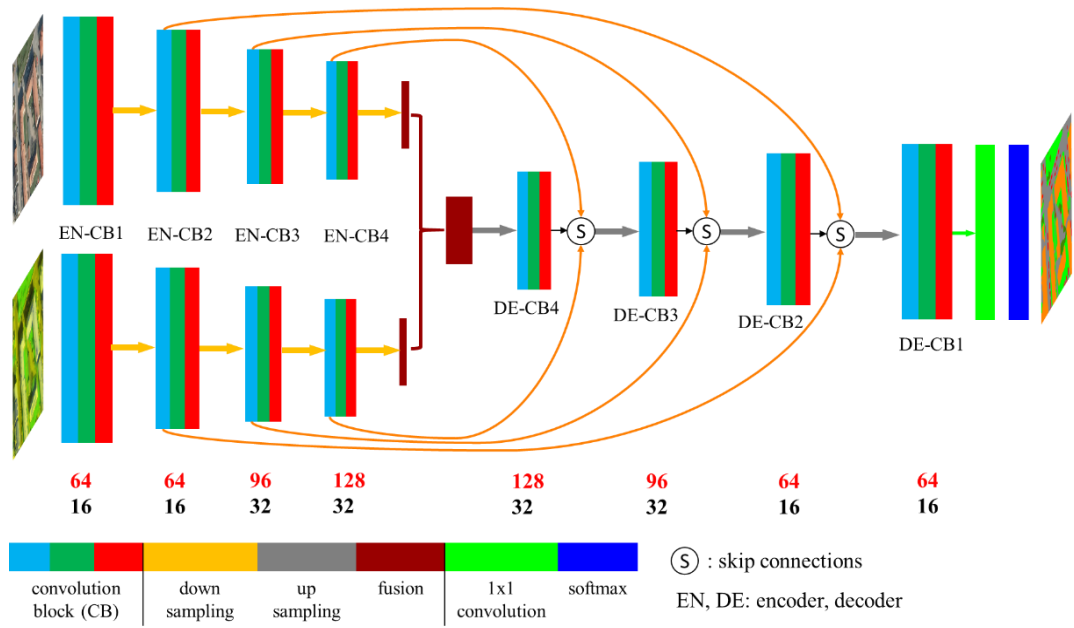

Figure 3: Architecture of *FuseNet* and *FuseNet-lite*. The numbers underneath the different convolutional blocks indicate the number of filters in the corresponding block, which is the only difference between the two networks: **red** numbers specify the number of filters for *FuseNet* whereas **black** ones are related to *FuseNet-lite*.

Both CNNs use an encoder-decoder architecture with skip-connections in which the encoder consists of two input branches, each of size 256 x 256 pixels with three bands. The entry of the first branch is an RGB orthophoto, whereas the entry of the second branch contains the height and the near infrared band of the orthophoto. While in principle the number of bands per entry can vary, in our experiments we use a three-band image consisting of the normalized DSM (nDSM, i.e. the difference between the DSM and the DTM) and the near infrared and red bands of the orthophoto. These two entries are processed independently by the two separate encoder branches, and the resultant features of the two branches are fused by 1 x 1 convolutions before decoding. Each block in the encoder consists of three convolution layers with filter size 3 x 3 and batch normalization (Ioffe et al., 2015) as well as a rectified linear unit (ReLU) after each convolution. At the end of each block, max-pooling is used to reduce the spatial resolution of the feature maps. Each block in the decoder upsamples the input by bilinear interpolation before forwarding it to three layers of 3 x 3 convolutions, batch normalization and ReLU. Except for the outmost convolution block, there are skip-connections between corresponding convolution blocks of the encoder and the decoder. These connections were found to be important to mitigate the problem of the loss of spatial information caused by the pooling operations (see Yang et al., 2020a).

The final 1 x 1 convolutional layer converts the output of the last convolutional layer to a vector of $M$ class scores for each of the $H \times W$ pixels of the input image, where $M$ is the number of land cover classes. This results in a vector $\mathbf{z}_{LC}^{i} = (z_{LC^1}^{i}, \ldots, z_{LC^M}^{i})^T$ of class scores for each pixel $i$ of the image, where $\mathbb{C}_{LC} = \{C_{LC^1}, \ldots, C_{LC^M}\}$ is the set of land cover classes and $z_{LC^c}^{i}$ is the class score for class $C_{LC^c}$. These class scores are normalized by a softmax function delivering the posterior probability $P_i(C_{LC^c}|X)$ for pixel $i$ to take class label $C_{LC^c}$ given the image data $x$:

$$P_i(C_{LC^c}|X) = \text{softmax}(\mathbf{z}_{LC}^{i}, C_{LC^c}) = \frac{exp(z_{LC^c}^{i})}{\sum_{m=1}^{M} exp(z_{LC^m}^{i})}, \qquad (1)$$

Training is based on stochastic mini-batch gradient descent (SGD) using backpropagation (see Section 4.2 for more details). We use the focal loss (Lin et al., 2017) as objective function for optimization, extended to be able to deal with multiple classes (see Yang et al., 2019 for details):

$$L_{LC} = -\frac{1}{W \cdot H \cdot N} \sum_{c,i,k} \left[ y_{LC^c}^{ik} \cdot (1 - P_i(C_{LC^c}|X_k))^{\gamma} \cdot log\big(P_i(C_{LC^c}|X_k)\big) \right], \qquad (2)$$

where $k$ is the index of an image, $X_k$ is the $k^{th}$ image in the mini-batch and $N$ is the number of images in a mini-batch. The variable $y_{LC^c}^{ik}$ takes the value of 1 if the training label of pixel $i$ in image $k$ is $C_{LC^c}$ and 0 otherwise, while the hyper-parameter $\gamma$, typically lying in the range of 1.0 to 5.0, controls the penalty term $(1 - P_i(C_{LC^c}|X_k))$. This term suppresses the influence of well-classified pixels (for which $(1 - P_i(C_{LC^c}|X_k))$ is close to zero), whereas it causes the loss to focus on falsely classified ones. Using this loss, it is possible to mitigate the problem of unbalanced class distributions, because classes for which a large number of samples is available are mostly well classified during training, so that this loss will put a higher emphasis on the samples of underrepresented classes (Lin et al., 2017).

### 3.1.2. Network modifications: *FuseNet-lite*

In the experiments conducted for (Yang et al., 2020a), we observed a tendency of *FuseNet* to overfit to the training data, indicated by a considerable gap between the training and validation errors. Due to the relatively large number of filters (red numbers in Fig. 3) and the depth of the network, the origi-nal network had a very large number of parameters (about 8.2 million). We attributed the tendency of *FuseNet* to overfit to an over-parameterization. In order to check this hypothesis, we gradually reduced the number of filters in each convolution block and checked the validation error. In this way, we found the network variant *FuseNet-lite* with a reduced number of filters (see the black numbers in Fig. 3), which only requires about 6% of the number of parameters of the original network while maintaining nearly the same level of classification accuracy. A comparison of the two networks, in particular with respect to potential overfitting, is given in Section 5.

## 3.2. Hierarchical land use classification
### 3.2.1. Overview

As pointed out at the beginning of Section 3, the goal of land use classification is to predict a set of class labels per database object, i.e. one per semantic level of the object catalogue, so that the predictions are consistent with the hierarchical object class catalogue. The first input consists of a land use database with a hierarchical object catalogue. In the database, objects are represented by polygons with land use information at multiple semantic levels. Furthermore, a multispectral orthophoto (RGB-IR), a nDSM and pixel-wise scores for land cover, e.g. delivered by *FuseNet-lite* (cf. Section 3.1), are required.

The classification is based on a CNN taking rasterized input data of a size of 256 x 256 pixels and returning multiple hierarchical labels simultaneously. In this context, the large variation of polygon size and shape is a challenge (see examples for a very long *road* and a small *settlement* object in Fig. 1), because the input size of the CNN is fixed. This means that large objects may not fit into the input window of the CNN, whereas very small objects might only cover a small percentage of the image at the geometrical resolution of the input data. To overcome the former problem, large objects not fitting into a window of 256 x 256 pixels are split into several tiles that are classified independently from each other. The way in which these tiles are prepared is described in section 3.2.2.

To overcome the problem of small polygons, Yang et al. (2019) proposed a CNN architecture in which the CNN is split into two branches before the classification layer where one branch contains the regular data, whereas the second branch uses an extract of the intermediate feature map that tightly encloses the object polygon. The resultant CNN structure, extended by a set of interacting classification heads for predicting class labels at multiple semantic levels simultaneously is described in section 3.2.3. A multi-task scenario for training this network is described in section 3.2.4. Although the mutual dependencies between the classes at different semantic levels are considered by links in the network and, thus, can in principle be learned from the data, the methods for inference and training described in sections 3.2.3 and 3.2.4, respectively, do not yet enforce semantic consistency; they serve as a baseline in our experiments. Section 3.2.5 describes how semantic consistency can be achieved in inference, whereas section 3.2.6 extends the training procedure by additional loss functions that should help the network to learn how to predict semantically consistent class labels. Finally, section 3.2.7 describes how the results are determined for objects that had to be split into multiple patches due to their size, which is achieved by merging the results of the individual tiles.

### 3.2.2. Preparation of input patches

Basically, the input data are prepared for the CNN by extracting a window of 256 x 256 pixels centred at the centre of gravity of the object from the input raster data. Additionally, a binary object mask representing the object shape is generated from the input polygon to be presented to the CNN. In this mask, a value of 1 indicates pixels inside the object, whereas all other pixels are set to 0. If the polygon fits into the window of 256 x 256 pixels at the GSD of the input image, the preparation is straightforward; if the object is very small, the window will be dominated by information outside the object, but this situation is not considered by a special strategy for data preparation; rather it is the motivation for a specific CNN structure (cf. Section 3.2.3). However, a problem arises if an object is larger than the input window, which is, for instance, frequently the case with road objects. Just scaling the object to fit into the window may reduce the GSD too much. Instead, the object is split into tiles that are classified independently. This strategy was found to be superior to scaling in (Yang et al., 2019).

In order to generate the input patches for the CNN, we split the window enclosing the entire database object into tiles of 256 x 256 pixels with an overlap of 50%. A tile is excluded from processing if the proportion of its area that overlaps with the object is small (less than 10% of the area of this tile). However, this might still lead to a large number of tiles for very large polygons. Thus, if more than $N_{min}$ tiles remain after this procedure, only 40% of them are selected randomly for being classified by the CNN (cf. Fig. 4) in order to reduce the computational burden. Otherwise, all of them are classified. We use $N_{min} = 3$ in our experiments.

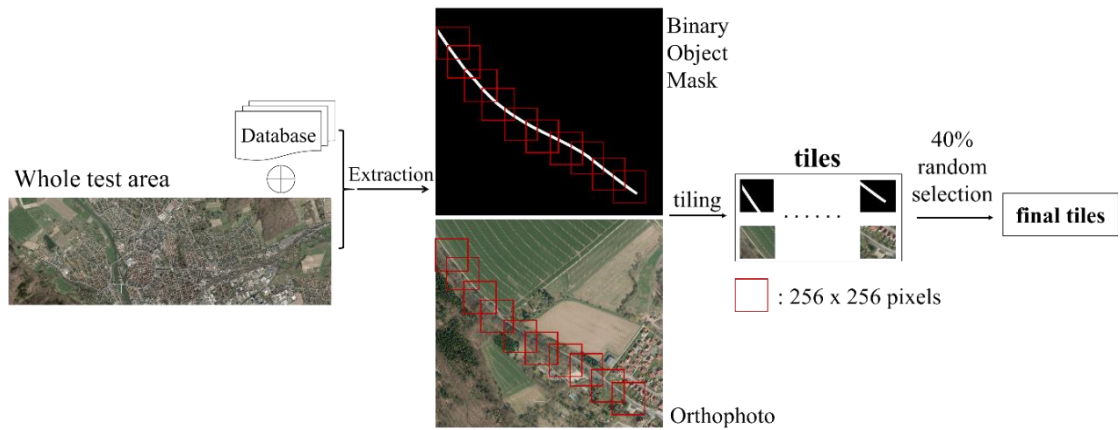

Figure 4: Example for tiling a large road object in one of the test datasets. Only the binary object mask representing the object shape and the RGB orthophoto are shown here to explain the principle, but the extracted tiles would contain all data mentioned in the main text.

For each tile thus selected, a raster image of 256 x 256 pixels consisting of $D = 6 + D_c$ bands is generated by combining the corresponding extracts from the binary object mask, the RGB-IR orthophoto, the nDSM (the first 6 bands) and the $D_c$ pixel-wise class scores from land cover classification (one per class) at the original GSD of the image data.

### 3.2.3. Basic CNN architectures: *LuNet-MT* and *LuNet-lite-MT*

The basic network architecture we use for land use classification is based on *LuNet* (Yang et al., 2019). Starting with several blocks of convolutional and pooling layers, the network is then split into two branches. The first of them continues with standard convolutional and pooling layers, while the second branch starts with the extraction of a region of interest (ROI) from the feature map of the previous joint layer that tightly encloses the object. The ROI contents are rescaled to 16 x 16 pixels; then, they are forwarded to a similar set of convolutions and pooling as in the first branch; cf. (Yang et al., 2019) for details. The resultant feature vectors of the two branches are concatenated and the combined vector is classified. As mentioned above, the ROI branch is mainly supposed to support the classification of very small objects. In preliminary experiments we noted a certain degree of overfitting, expressed by a relatively large discrepancy between training and test errors. Therefore, for this paper we shrink the architecture by reducing the number of filters to be learned in a way similar to what has been described for land cover in Section 3.1, leading to the network variant *LuNet-lite*. Fig. 5 shows both network variants; the number of filters in each convolution block are shown in red for the original *LuNet*, while the corresponding number of filters in *LuNet-lite* is shown in black. Compared to *LuNet*, which has about 14 million parameters, the number of parameters of *LuNet-lite* is reduced to 7%.

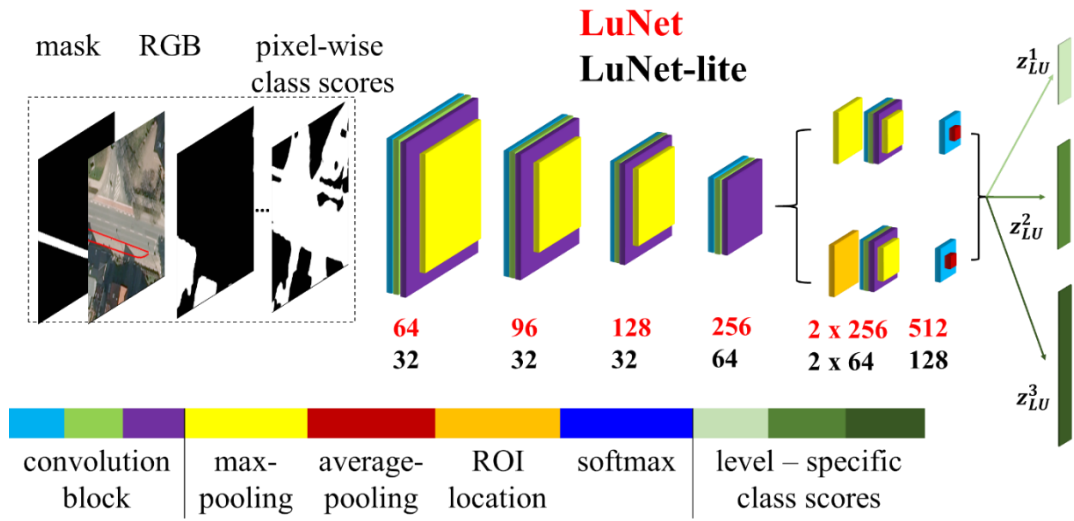

Figure 5: Basic architecture of *LuNet* and *LuNet-lite* for B = 3 semantic levels (level I / coarsest level – level III / finest level). The numbers underneath the convolution blocks shows the number of filters in the corresponding convolution block: **red** numbers are for *LuNet* whereas **black** numbers refer to *LuNet-lite*. The integration of the semantic dependencies is shown in detail in Fig. 6. For the explanation of the mathematical symbols, refer to the main text.

Fig. 6 shows the detailed network structure with three interacting classification layers to predict one label per level and to integrate the semantic dependencies between the classes at different levels for the three levels. It represents a multi-task classification scenario, with the tasks corresponding to the predictions of the labels at different semantic levels. Whereas the class labels are predicted by independent classification heads, the prediction is based on a shared representation determined by the earlier CNN layers (consisting of 512 and 128 features for *LuNet* and *LuNet-lite*, respectively).

In Figures 5 and 6, roman numerals indicate the semantic levels. For each semantic level $l$, the concatenated feature vector from the two branches of the network is passed through one fully connected (FC) layer that delivers a vector of un-normalized class scores $\mathbf{z}_{LU}^{l} = \left( z_{LU,1}^{l}, \ldots, z_{LU,M_l}^{l} \right)^{T}$ for the $M_l$ classes $C_{LU,c}^{l} \in \{C_{LU,1}^{l}, \ldots, C_{LU,M_l}^{l}\}$ of the set of land use classes at category level $l$, where $z_{LU,c}^{l}$ is the class score predicted for class $C_{LU,c}^{l}$ on the basis of an image $X$. These vectors $\mathbf{z}_{LU}^{l}$ form the input for the multi-task classification head, which applies a specific information flow for learning the semantic relationships (Fig. 6). In the first layer, information flows from the coarser to the finer levels, whereas in the second layer, the information flow is reversed. In the remainder of this paper, the expanded network is referred to as *LuNet-MT* or *LuNet-lite-MT* (MT for multi-task), depending on whether it is built on top of *LuNet* or *LuNet-lite*, respectively. The structure of *LuNet-MT* is identical to the one described in (Yang et al., 2020b), but we use a simpler (though equivalent) mathematical description of the way in which the classification head works. It is inspired by (Hu et al. 2016), though embedded in a completely different context and expands the degree of interaction between the individual semantic levels.

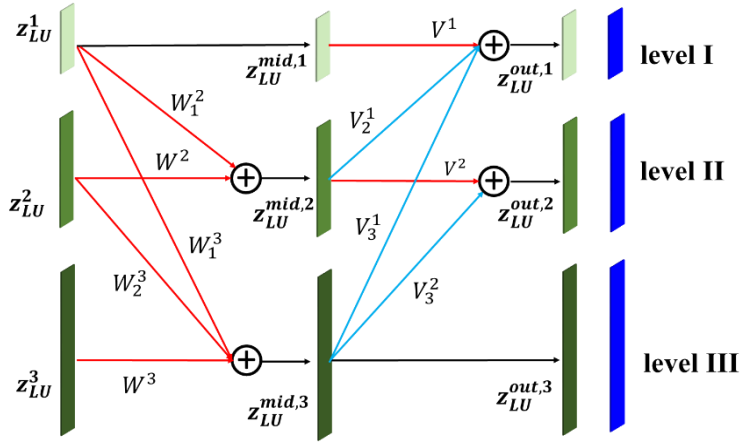

Figure 6: Architecture of the network for integrating the semantic dependencies, showing the details of the last processing block in Fig. 5 for $B=3$ semantic levels (from I … coarsest to III … finest level). For the explanation of the mathematical symbols, refer to the main text.

The first layer of the classification head determines intermediate class scores $\mathbf{z}_{LU}^{mid,l}$ at each level $l$. For $l > 1$, they are based on the output of the previous network layer for all coarser levels and for the same one; otherwise (for $l = 1$), the scores are just copied:

$$\mathbf{z}_{LU}^{mid,l} = \begin{cases} W^l \cdot ReLU(\mathbf{z}_{LU}^{l}) + \sum_{i=1}^{l-1} W_i^l \cdot ReLU(\mathbf{z}_{LU}^{i}), & if\ l > 1 \\ \mathbf{z}_{LU}^{l}, & if\ l = 1 \end{cases}. \quad (3)$$

In eq. 3, $W^l$ and $W_i^l$ are the parameters to be learned for that layer.

The second layer of the classification head determines the final un-normalized class scores $\mathbf{z}_{LU}^{out,l}$.

Whereas for the finest level ($l = B$), these scores are just copies of the intermediate ones, the others receive input from all finer levels and from the same one ($l < B$):

$$\mathbf{z}_{LU}^{out,l} = \begin{cases} V^l \cdot ReLU(\mathbf{z}_{LU}^{mid,l}) + \sum_{j=l+1}^{B} V_j^l \cdot ReLU(\mathbf{z}_{LU}^{mid,j}), & if\ l < B \\ \mathbf{z}_{LU}^{mid,l}, & if\ l = B \end{cases}. \quad (4)$$

In eq. 4, $V^l$ and $V_j^l$ are the parameters of that layer. By learning the parameters of these two FC layers, the network is supposed to learn the semantic relations between classes at different semantic levels, though no measures are taken to enforce consistency with the hierarchical object catalogue.

The class scores $\mathbf{z}_{LU}^{out,l}$ resulting from eq. 4 are normalized by a softmax function to obtain values that can be interpreted as probabilities:

$$P(C_{LU,c}^l|X) = softmax(\mathbf{z}_{LU}^{out,l}, C_{LU,c}^l) = \frac{exp(z_{LU,c}^{out,l})}{\sum_{m=1}^{M_l} exp(z_{LU,m}^{out,l})}. \quad (5)$$

Note that in this multi-task scenario, dependencies between the class labels are considered by the architecture of the classification heads. However, there is no guarantee that the class labels that are obtained by maximising eq. 5 will be semantically consistent. The modification of the inference architecture to guarantee such consistency is described in section 3.2.5.

### 3.2.4. *LuNet-MT* and *LuNet-lite-MT*: Training

We train these networks using stochastic mini-batch gradient descent (SGD) with weight decay, momentum and step learning policy. SGD is used to minimize the multi-class extension of the focal loss (Yang et al., 2019), which was found to converge faster than the cross-entropy loss in our previous work and which is also supposed to mitigate the problems related to unbalanced class distributions of the training data:

$$L_{LU} = -\frac{1}{N} \cdot \sum_{l,c,k} \left[ y_c^{l,k} \cdot (1 - P(C_{LU,c}^l|X_k))^\epsilon \cdot log\left(P(C_{LU,c}^l|X_k)\right) \right]. \quad (6)$$

In eq. 6, $X_k$ is the $k^{th}$ image in a mini-batch, $N$ is the number of images in a mini-batch, and $y_c^{l,k}$ is 1 if the training label of $X_k$ is $C_{LU,c}^l$ in level $l$ and 0 otherwise. The parameter $\epsilon$ is the hyper-parameter to control the penalty term (corresponding to $\gamma$ in eq. 2).

### 3.2.5. Achieving semantic consistency between the individual levels

As pointed out at the end of section 3.2.3, *LuNet-MT* and *LuNet-lite-MT* deliver predictions of multiple class labels corresponding to different semantic levels simultaneously, while considering semantic dependencies as they were learned from the training samples. However, there is no guarantee for the predictions to be consistent with the object catalogue hierarchy. In our previous work, we proposed two strategies to enforce consistency with the class hierarchy, both of them based on greedy sequential post-processing of the class scores delivered by the multi-task CNN. In this paper, we adopt the better-performing strategy (*F2C; fine-to-coarse*, see Yang et al. 2020b for details) as a baseline and compare it to a newly proposed strategy (*JO; joint optimization*) to obtain the most probable set of class labels that is consistent with the class hierarchy based on the output of the multi-task head. The *JO* strategy is illustrated in Fig. 7.

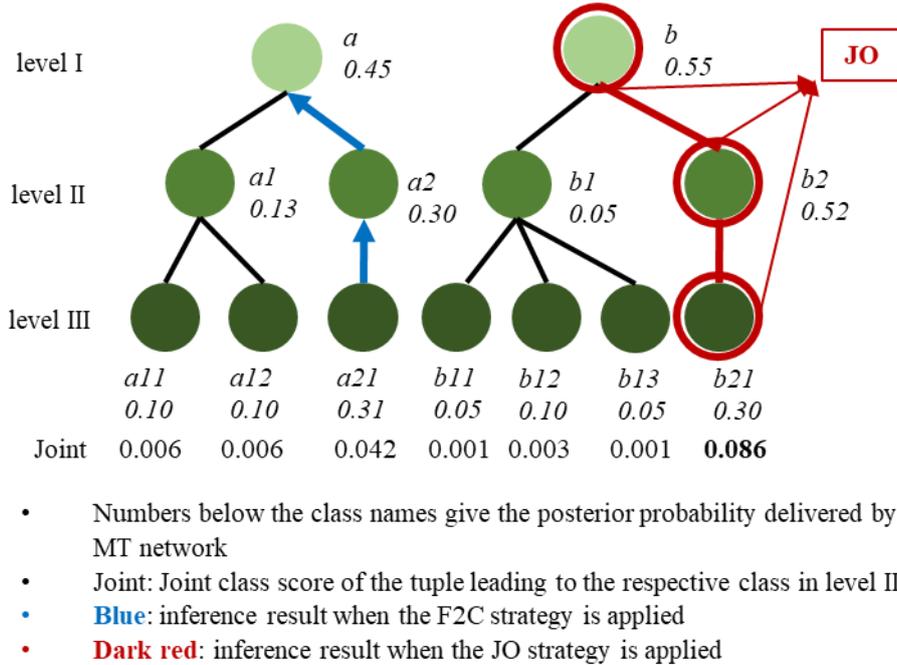

- Numbers below the class names give the posterior probability delivered by the MT network
- Joint: Joint class score of the tuple leading to the respective class in level III
- **Blue**: inference result when the F2C strategy is applied
- **Dark red**: inference result when the JO strategy is applied

Figure 7: Illustration of the *JO* strategy using an example in which labels are predicted at three semantic levels. The lines indicate hierarchical relations between classes. Strategy *JO* selects the tuple of class labels consistent with the hierarchy that has the highest joint class score (dark red). Strategy *F2C* starts the prediction at the finest level and use it to control predictions at coarser levels (blue).

In the *F2C* strategy, we make the first prediction at the finest level by selecting the class label which has the maximum class probability, and then use it to control the predictions at the coarser levels along with the class hierarchy. One drawback of *F2C* strategy is this sequential processing: if the first prediction made is wrong, this error cannot be corrected at a later stage.

Using the *JO* strategy, we obtain the most probable set of class labels consistent with the class hierarchy. Similar to the multitask-setting (cf. section 3.2.3), we start by predicting probabilistic scores for each level according to eq. 5. We then compute combined class scores for all possible combinations of classes at different semantic levels. Given the hierarchical class structure, a set of class labels that is consistent with that class structure is a *B*-tuple of class labels (a triple in our case) that corresponds to all labels along a path through the tree representing the class structure for one of the classes of level I (cf. the dark red path in Fig. 7). There are $M_B$ valid tuples $T_i = \{C^1_{LU,i}, C^2_{LU,i} C^3_{LU,i}\}$, one per class label differentiated at the finest level of the hierarchy, where $C^l_{LU,i}$ is the class label of the $i^{th}$ tuple at level $l$. For each input image $X_k$ and each tuple $T_i$, we compute the joint class scores for the image to correspond to that tuple according to:

$$P_{joint}^{i,k}(T_i, X_k) = \prod_{l=1}^{B} P(C_{LU,i}^l | X_k), \qquad (7)$$

where $P(C_{LU,i}^l | X_k)$ is the output of the softmax function (eq. 5) for class $C_{LU,i}^l$ at the semantic level $l$. The classification output for an image $X_k$ is the tuple $T_i$ that maximises the joint class scores according to eq. 7.

In our experiments, we compare two different CNN variants based on this strategy. They are referred to as *LuNet-JO* and *LuNet-lite-JO* (the abbreviation MT is omitted for simplicity in the remainder) and apply the joint inference strategy described in this section on the basis of the *LuNet-MT* and *LuNet-lite-MT* architectures, respectively, both described in section 3.2.3.

### 3.2.6. Considering the class hierarchy in training

In this section, we present a training strategy based on the joint optimization described in the previous section that also considers the class hierarchy in the loss function. In this training procedure, we minimize a loss function $L_{LU}^{JO}$ consisting of two components $L_{LU}^{JO,P1}$ and $L_{LU}^{JO,P2}$:

$$L_{LU}^{JO} = L_{LU}^{JO,P1} + L_{LU}^{JO,P2} \qquad (8)$$

In order to consider the constraints imposed on the class labels by the hierarchical object catalogue, we maximize the joint class scores $P_{joint}^{i,k}(T_i, X_k)$ (eq. 7) of the hierarchical tuples matching the ground truth, which is considered by the first term in eq. 8:

$$L_{LU}^{JO,P1} = -\frac{1}{N} \cdot \sum_k \sum_i^{M_B} y_i^{B,k} \cdot (1 - P_{joint}^{i,k}(T_i, X_k))^\epsilon \cdot \log\left(P_{joint}^{i,k}(T_i, X_k)\right). \qquad (9)$$

At the same time, we want to minimize the joint class scores of the incorrect tuples, which leads to the second loss function term:

$$L_{LU}^{JO,P2} = -\frac{1}{N} \cdot \sum_k \sum_i^{M_B} (1 - y_i^{B,k}) \cdot (P_{joint}^{i,k}(T_i, X_k))^\epsilon \cdot \log(1 - P_{joint}^{i,k}(T_i, X_k)). \qquad (10)$$

In eqs. 9 and 10, $y_i^{B,k}$ is the ground truth label of image $X_k$ at the finest level $B$ which, thus, identifies the correct tuple $T_i$, and $N$ denotes the number of images in a mini batch. We also introduce the focal-loss-style penalty terms as in eq. 6. This strategy for training is applied to the variants involving joint optimization, i.e. *LuNet-JO* and *LuNet-lite-JO* (cf. section 3.2.5).

### 3.2.7. Inference at object level

The network variants described in sections 3.2.3 and 3.2.5 are used to predict class labels for an input image patch of 256 x 256 pixels. The inference of the land use class labels is straight-forward for objects that fit into such a window and which are, thus, not split into tiles: the prediction of CNN for the image patches containing the object is the final result. The inference of objects which had to be split differs between the network variants:

- Variants *LuNet-MT* and *LuNet-lite-MT* (section 3.2.3): For every patch generated in the way described in section 3.2.2, we obtain class scores, and the product of the class scores from all patches is used to select the most probable class for the compound object at each semantic level.

- Variants *LuNet-F2C* and *LuNet-lite-F2C* (section 3.2.5): Here, the decision about the class labels at the finest level (III) is based on a majority vote of the results for the individual patches. The labels at the coarser semantic levels are determined in a straightforward way by selecting the ancestor categories of the predictions at the finest level.

- Variants *LuNet-JO* and *LuNet-lite-JO* (section 3.2.5): Here, we start with a combination of the class scores of the individual patches in the same way as described for *LuNet-MT*. This results in a vector of combined probabilistic class scores for every class at every semantic level. Afterwards, we compute the joint class scores of all consistent tuples (eq. 7) using the combined class scores, and we select the tuple that maximises this joint class score.

## 4. Test data and test setup

### 4.1. Test data

#### 4.1.1. Hameln and Schleswig

The main dataset used in our experiments and the one we use for evaluating both land cover and land use classification consists of two test sites located in Germany. The first one, Hameln, covers an area of 2 x 6 km$^2$ and shows various urban and rural characteristics. The second test site, Schleswig, covers an area of 6 x 6 km$^2$ and has similar characteristics. The input consists of digital orthophotos (DOP), a DTM, and a DSM derived by image matching. Subtracting the DTM from the DSM resulted in a normalized DSM (nDSM), which served as the input height channel in our experiments. In addition, land use objects from the German Authoritative Real Estate Cadastre Information System (ALKIS) are available, so that ALKIS serves as the geospatial database. The DOP are multispectral images (RGB + near Infrared / IR) with a GSD of 20 cm. There is a domain gap between these two datasets mainly caused by the different seasons (spring and summer for Hameln and Schleswig, respectively). The reference for land cover consists of 37 and 26 image patches (1000 x 1000 pixels each) for Hameln and Schleswig, respectively. The reference, generated by manual labelling, distinguishes 8 land cover classes: *building (build.), sealed area (seal.), bare soil (soil), grass, tree, water, car* and *others*. The reference for land use was derived from ALKIS, considering the first three semantic levels of its hierarchical object catalogue (AdV, 2008). Some object classes at levels II and III which exhibited a very low number of incidences only were combined. The details of the resulting class structure along with the number of samples are presented in Tab. 1. There are 4 classes in level I, 14 classes in level II and 21 classes in level III.

| level I | level II | level III | # Hameln | # Schleswig |
|---|---|---|---|---|
| settlement | residential (res.) | residential in use (res.use) | 528 | 803 |
| | | extended residential (ext. res.) | 34 | 61 |
| | industry (ind.) | factory (fact.) | 87 | 39 |
| | | business (busi.) | 193 | 158 |
| | | infrastructure (infra.) | 54 | 62 |
| | mixed usage (mix) | mixed usage (mix) | 9 | 127 |
| | special usage (special) | special usage (special) | 135 | 207 |
| | recreation (rec.) | leisure (leis.) | 27 | 64 |
| | | park | 299 | 365 |
| traffic | road traffic (ro.traf.) | motor road (mo.road) | 530 | 732 |
| | | traffic guided area (traf.area) | 134 | 75 |
| | path & way (path) | path & way (path) | 477 | 287 |
| | parking lot (park.lot) | parking lot (park.lot) | 91 | 76 |
| vegetation | agriculture (agr.) | farm land (farm) | 58 | 214 |
| | | garden / fallow land (garden) | 100 | 440 |
| | forest | hardwood or softwood (h/s.wood) | 33 | 154 |
| | | hardwood and softwood (h&s.wood) | 15 | 134 |
| | grove | grove | 51 | 88 |
| | moor or swamp (moor) | moor or swamp (moor) | 31 | 116 |
| water bodies | flowing water (flow.wat.) | flowing water bodies (flow.wat.bo) | 54 | 41 |
| | stagnant water (stag.wat.) | stagnant water bodies (stag.wat.bo.) | 5 | 102 |
| | | Total number of objects | 2945 | 4345 |

Table 1: Hierarchical land use class structure and statistics about the distribution of objects in Hameln and Schleswig. The class labels are given at three semantic levels (I-III) according to (AdV, 2008). Abbreviations that are used in the remainder of this paper are shown in brackets. The two rightmost columns give the number of land use objects of the given category in level III.

### 4.1.2. Vaihingen and Potsdam

In order to make our results for land cover classification comparable to the state of the art, we use the Vaihingen and Potsdam datasets from the ISPRS 2D semantic labelling challenge (Wegner et al., 2017). For Vaihingen, 33 colour infrared (CIR) images (each about 2000 x 2500 pixels) with a GSD of 9 cm are available, whereas in Potsdam, there are 38 orthophotos (RGB-IR) (each 6000 x 6000 pixels) with a GSD of 5 cm. In addition, nDSM are provided by Gerke (2015). For both datasets, six land cover classes are to be discerned: *impervious surface (imp. surf.), building (build.), low vegetation (low veg.), tree, car* and *clutter* (Wegner et al., 2017).

### 4.2. Test setup for land cover classification

Our implementation of the methods for pixel-based classification of land cover described in section 3.1 is based on TensorFlow (Abadi et al., 2015). We use a GPU (Nvidia TitanX, 12GB) to accelerate training and inference. To train *FuseNet-lite* and *FuseNet*, we applied SGD with momentum of 0.999, minimizing the loss in eq. 2 along with a regularization term based on weight decay (Bishop, 2006) with a weight parameter of 0.0005. The hyper-parameter of focal loss in eq. 2 was set to $\gamma = 1$. The learning rate was set to 0.1 and decreased to 0.01 after 15 epochs in a total of 30 epochs training. The mini-batch size was set to 10 for *FuseNet-lite* and 4 for *FuseNet*. We could use a larger mini-batch size for *FuseNet-lite* due to the smaller memory footprint of the network, which allowed us to keep more training images in memory.

### 4.2.1. Test setup for Hameln and Schleswig

The CNN variants for the pixel-based prediction of land cover described in section 3.1 (*FuseNet* and *FuseNet-lite*) require a RGB image and a second image of three bands that is the combination of the red and the infrared bands of the orthophoto and the nDSM. All images (RGB, RID and reference label images) were cut into four non-overlapping tiles of size 500 x 500 pixels, resulting in 148 and 104 tiles for Hameln and Schleswig, respectively. These tiles were randomly split into three groups of equal size for three-fold cross validation. Each tile was then split into four partly overlapping patches corresponding to the input size of the CNN (256 x 256 pixels). In each test run, one group of tiles was used for testing and the others are used for training. During training, about 10% of the samples of the training set were reserved for validation. For both sites, we applied data augmentation by rotations of 90°, 180°, 270°, as well as horizontal and vertical flipping during training. Additionally, two random rotations in the range of [3°,20°] were applied to all patches available after the first data augmentation step, so that the total number of patches available for training was 18 times the original one. We report average quality metrics derived from the confusion matrices determined on a per-pixel level in all test runs. In particular, we present the overall accuracy (OA), the class-specific F1 scores, i.e. the harmonic mean of completeness and correctness, and the mean F1 score (mF1), i.e. the average of the F1 scores of all classes. We consider mF1 to be an overall quality measure that is more susceptible to problems with underrepresented classes than OA.

We pursue two goals of the experiments for evaluating the CNN for land cover in Hameln and Schleswig. Firstly, we want to compare the two network variants described in section 3.1 to see if the reduction of the number of parameters reduces overfitting without a large drop in classification performance. In order to do so, we train and test the network variants using only data from a single test site and compare the results. The second goal is to assess whether the availability of a larger set of training data is beneficial even if there is a certain domain gap between different subsets of the training and test samples. This is only evaluated for *FuseNet-lite*. In order to do so, we generate a combined set of training and test samples involving both datasets and repeat the computations, again based on three-fold cross validation. Here we take care that in each group there is a mixture of training and test samples from both datasets. The experiments related to land cover are presented in section 5.1.

### 4.2.2. Test setup for Vaihingen and Potsdam

These benchmark datasets were applied to *FuseNet-lite* to allow a comparison of this classification method to the state of the art. Following the benchmark protocol, in Vaihingen 16 images with known reference were used for training and the rest (17) for testing, while in Potsdam 24 images with known reference were used for training and the rest (14) for testing. In Vaihingen, four of the training images (Image IDs 5, 7, 23, 30) were used for validation and only the remaining 12 images were used to determine the parameters of the networks. In Potsdam, the validation set consisted of three of the training images (Image IDs 02-11, 06-10, 04-10) and the rest (21) was used to determine the parameters. As *FuseNet-lite* requires an input of 256 x 256 pixels, windows of 256 x 256 pixels with an overlap of 128 pixels in both spatial dimensions were extracted from the training images, resulting in 3328 training patches for Vaihingen and 44436 training patches for Potsdam, respectively. Data augmentation based on rotations of 90°, 180°, 270°, horizontal and vertical flipping was applied for both test sites, thus multiplying the number of available training patches by a factor of six. As the resultant number of training patches for Vaihingen was relatively small compared to Potsdam, we followed the additional augmentation strategy applied for Hameln and Schleswig for the Vaihingen dataset (cf. Section 4.2.1), that is, we applied two additional random rotations in the range of [3°,20°] to the patches obtained after the first augmentation step, thus tripling the number of training patches. In Vaihingen, due to the lack of a blue band, we use CIR instead of RGB images as the first input and a composite of the red and near infrared bands and the nDSM (RID) as the second input. In Potsdam, RGB and the composite RID served as the inputs. During inference, the class labels for a patch of 256 x 256 pixels were predicted six times for the original image and variants that were flipped and rotated as the training images, and the probabilistic scores were multiplied to obtain a combined score for classification.

The final evaluations were based on the test images (17 images in Vaihingen and 14 images in Potsdam; cf. Section 4.1.2). The reported evaluation metrics are again the OA, class-specific F1 scores and mean F1 score over all classes. Following the benchmark protocol, we primarily used the eroded reference to obtain these quality metrics, which does not consider pixels near object boundaries and which was generated by an erosion of the full reference with a circular disc of a radius of 3 pixels. However, metrics based on the full reference are also provided.

### 4.3. Test setup for land use

The networks for land use classification described in section 3.2 are also implemented based on the TensorFlow framework (Abadi et al., 2015), and the hardware used for the experiments related to land cover was also used to evaluate the land use classification.

In all experiments related to land use classification, we use stochastic mini-batch gradient descent with momentum of 0.999 for minimizing the loss functions mentioned in the previous sections. For the training of all network variants, the hyper-parameter of the focal loss (eq. 6, 9, 10) is set to $\epsilon = 1$, and the hyper-parameter for weight decay is 0.0005. The setting of these hyper-parameters follows the setting in our previous work (Yang et al., 2019), where we found it to deliver good results. All networks are trained for eight epochs, where in one epoch all samples are used for training in one iteration. In this context, we use a base learning rate of 0.001 and reduce it to 0.0001 after four epochs. The mini-batch size is set to 18 for *LuNet*-based variants and to 30 for *LuNet-lite*-based variants. The reason that we can use larger mini-batch sizes for the variants based on *LuNet-lite* is again the smaller memory footprint of that network.

Only the Hameln and Schleswig datasets described in section 4.1.1. were used in this set of experiments; we did not find a benchmark dataset for land use classification that fulfilled the requirements for testing our methodology (i.e., database polygons and annotations according to a hierarchical class structure). Each of the two datasets is split into six blocks for cross validation. The block size is 10.000 x 5.000 pixels (2 km$^2$) and 30.000 x 5.000 pixels (6 km$^2$) for Hameln and Schleswig, respectively. The number of land use objects and the class structure are shown in Tab. 1. In each test run, five blocks are used for training and one for testing. In each run, about 15% of the training samples (i.e., database objects) are again used for validation, and the rest is used for determining the network parameters. There is no overlap between the training, validation and test sets. Data augmentation is applied in both sites. We select one random rotation for each interval of 30° width for polygons that have to be split into tiles (cf. section 3.2.2), while we select one random rotation for each interval of 5° width for all other polygons. There are different parameters for large polygons (those having to be split) and small ones because in our datasets the number of tiles of large polygons is about six times larger than the number of small polygons. As a result, we obtain 354.178 and 479.978 patches for Hameln and Schleswig, respectively. As there are eight land cover classes (cf. section 4.1), each patch has $D = 14$ bands. The results of *FuseNet-lite* are used to generate the class scores for land cover. We compare all network variants described in sections 3.2.3 and 3.2.5. The evaluation is based on the number of correctly classified database objects. We report the average overall accuracy (OA) and F1 scores over all test runs of cross validation. To investigate whether adding data of different characteristics could help classification or not, we mix both datasets to form a new 6-fold cross validation setup by combining the samples of the blocks indicated by the same block number. As land cover classification is more directly affected by the physical appearance of the pixels in the image than land use, we still use individual training for the CNN used to derive the land cover class scores in the combined land use classification scenario.

## 5. Experimental Evaluation

In this section, we describe and analyse the results of the experimental validation of our methodology using the data and strategies described in section 4. The experiments related to land cover are presented in section 5.1, while section 5.2 is dedicated to the evaluation of land use classification.

### 5.1. Evaluation of land cover classification
#### 5.1.1. Hameln and Schleswig

Table 2 presents the results of land cover classification for the two CNN variants *FuseNet* and *FuseNet-lite* when trained and tested on one dataset only. In general, both networks show a good performance with overall accuracies between 86% and 90% in all cases. A comparison of all variants shows that *FuseNet* trained on individual datasets delivers the best results (OA of 89.8% and 87.2% in Hameln and Schleswig, respectively). However, the margin in the OA between the two variants when trained on individual datasets is relatively small (1% and 0.7% in Hameln and Schleswig, respectively). It would seem that even though it only requires 10% of the parameters of *FuseNet*, *FuseNet-lite* performs almost equally well. Somewhat larger difference can only be observed in the mean F1 scores (2.6% / 3.8% in Hameln and Schleswig, respectively), which is mainly caused by the reduced F1 scores of the classes *car* and *others* (7.2% / 8.1% in Hameln and 8.6% / 18.2% in Schleswig). These two classes are underrepresented, having fewer training samples than other classes. For instance, in Hameln, the number of samples of class *car* is about 2% of the whole training set, and about 1.3% belong to the class *others*. The two classes are irrelevant for land use classification, given our application of the verification of geospatial databases.

| Test site | Network | F1 [%] | | | | | | | | mF1 [%] | OA [%] |
|---|---|---|---|---|---|---|---|---|---|---|---|
| | | *build.* | *seal.* | *soil* | *grass* | *tree* | *water* | *car* | *others* | | |
| Hameln | *FuseNet* | **95.1** | **88.7** | **83.5** | **89.1** | **89.3** | **95.9** | **79.4** | **50.4** | **83.9** | **89.8** |
| | *FuseNet-lite* | 94.2 | 87.0 | 82.8 | 88.2 | 88.7 | 95.2 | 72.2 | 42.3 | 81.3 | 88.8 |
| Schleswig | *FuseNet* | **92.4** | **85.1** | **78.7** | **84.2** | **91.6** | 91.2 | **69.4** | **41.9** | **79.3** | **87.2** |
| | *FuseNet-lite* | 92.2 | 84.2 | 77.6 | 83.1 | 91.3 | **91.3** | 60.8 | 23.7 | 75.5 | 86.5 |

Table 2: Results of the evaluation of land cover classification when training and testing on individual test sites. mF1: mean F1 score; OA: overall accuracy.

Fig. 8 shows a comparison of the training and validation errors, exemplarily for the three test runs in Hameln. For *FuseNet*, the difference between the training and validation errors is in the order of 7%, which we take as an indicator that the network overfits. For *FuseNet-lite*, overfitting is reduced by about 30%, although the validation error is nearly unchanged and the OA is only reduced by 1% (cf. Tab.2). Thus, we use the results of *FuseNet-lite* for further processing due to its much smaller memory footprint and because it is less susceptible to overfitting.

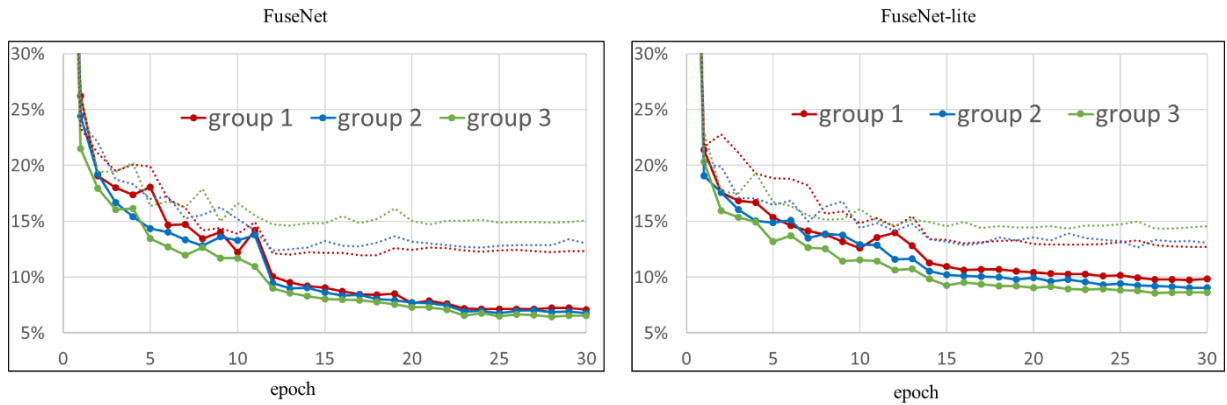

Figure 8: Training and validation errors of land cover classification in Hameln as a function of the training epoch. Solid lines: training error; dashed lines: validation error. Errors are given as the rate of incorrectly classified pixels. They are given separately for each of the three test runs of cross validation (groups 1 – 3).

Table 3 shows the results of the evaluation of land cover classification achieved by *FuseNet-lite* when combining training and test samples from the two datasets. The classification results in Schleswig are improved by +0.7% and +2% in terms of OA and mean F1 score, respectively. In Hameln, there is only a slight improvement of +0.3% in terms of mean F1 score. Considering individual classes, in Hameln most classes are slightly better recognized except *grass* and *water*, which have drops of F1 scores of 0.3% and 1.4%, respectively. Similar improvements can be observed in Schleswig, except that the F1 score of class *water* drops by 1.9%. In both sites, the decrease in the performance for *water* in the combined dataset may be due to the different appearance of that class in Hameln and Schleswig, respectively. The lower half of Fig. 9 shows examples of *water* inside forest areas in both sites. In

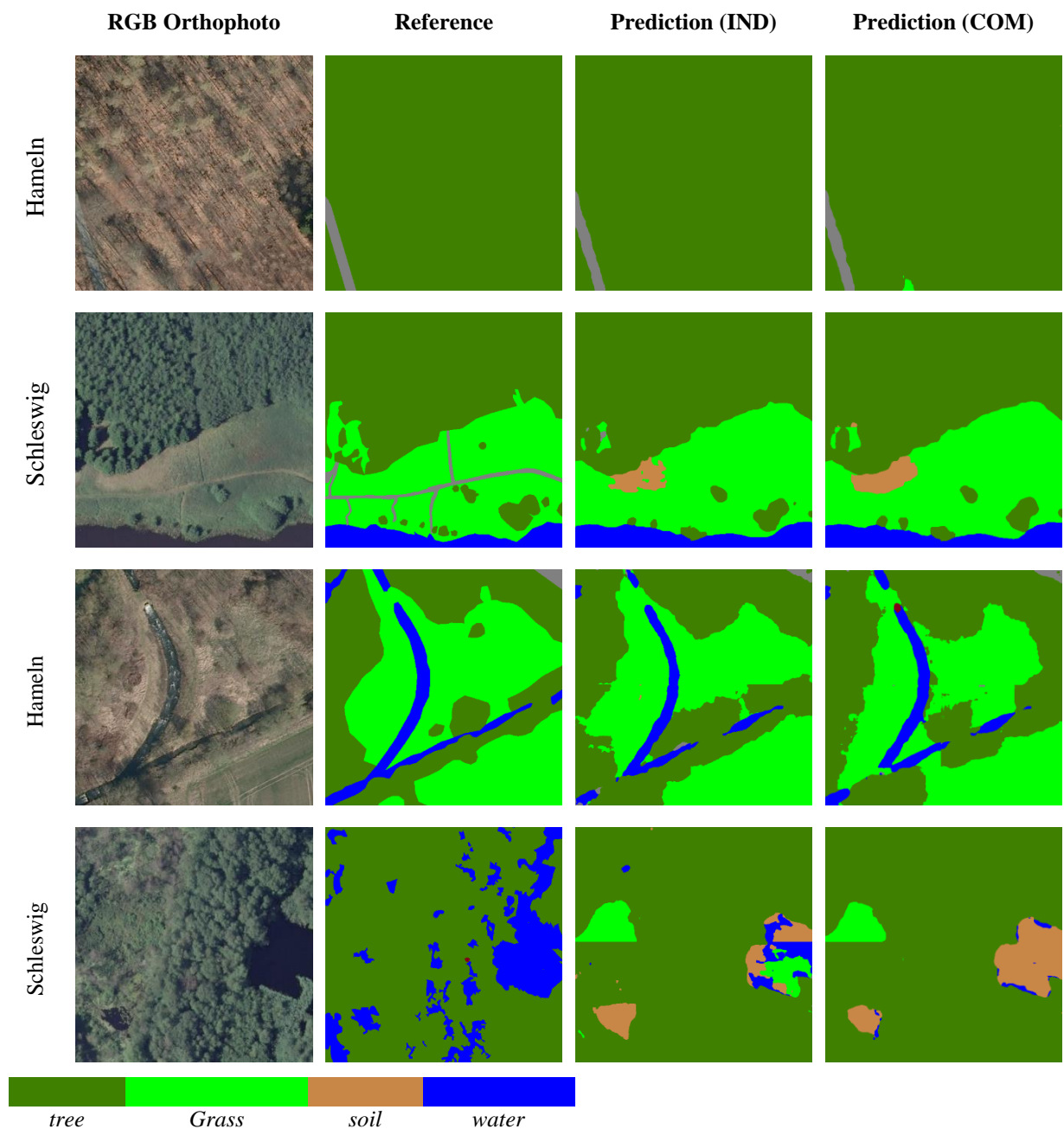

Figure 9: Examples for the different visual appearance of *tree* and *water* in Hameln and Schleswig, respectively, and corresponding classification results of *FuseNet-lite*. The figure shows the orthophoto, the land cover reference and the results of land cover classification. Prediction (IND): the network was trained only using samples from the considered test site; Prediction (COM): the network was trained using a combination of samples from both test sites.

Schleswig, water pixels are affected by the shadow of trees, whereas in Hameln they are exposed clearly. This improvement of the results by joint training are in line with the findings of Kaiser et al. (2017), who found that using a large amount of data with inaccurate label information (the latter derived from Open Street Map) could improve the performance of land cover classification if some high-quality labels for the area to be classified are available. Although there is large difference in the appearance of vegetation objects (i.e. *grass*, *tree*), their separability is not affected too much after combining the data from the two test sites. The upper half of Fig. 9 shows two examples involving trees indicating that training based on the combined dataset delivers similar results as individual training.

| Test site | F1 [%] | | | | | | | mF1 [%] | OA [%] |
|---|---|---|---|---|---|---|---|---|---|
| | *build.* | *seal.* | *soil* | *grass* | *tree* | *water* | *car* | *others* | |
| Hameln | 94.3 | 87.2 | 83.4 | 87.9 | 88.7 | 93.8 | 72.2 | 45.4 | 81.6 | 88.8 |
| Schleswig | 92.1 | 84.4 | 78.3 | 83.9 | 91.3 | 89.4 | 60.7 | 40.0 | 77.5 | 87.2 |
| Hameln + Schleswig | 93.7 | 86.7 | 81.2 | 87.0 | 90.1 | 92.5 | 73.3 | 43.8 | 81.0 | 88.4 |

Table 3: Results of land cover classification *FuseNet-lite* based on a training set that is a mixture of samples from the two test sites. mF1: mean F1 score; OA: overall accuracy.

### 5.1.2. Comparison to the state-of-the-art: Vaihingen and Potsdam

Tab. 4 presents a comparison of the results achieved for Vaihingen and Potsdam by *FuseNet-lite* with those achieved by other methods. The comparison considers the scoreboard of the ISPRS benchmark (Wegner et al., 2017), but also other recent publications (the scoreboard has not been updated anymore since 2018).

| Network | Input | #Param. | F1 [%] | | | | | mF1 [%] | OA [%] |
|---|---|---|---|---|---|---|---|---|---|
| | | | *build.* | *imp. surf.* | *low veg.* | *high veg.* | *car* | | |
| **Vaihingen** | | | | | | | | | |
| HUSTW (Sun et al., 2019) | CIR, nDSM | $1.3 \cdot 10^7$ | 96.1 | 93.3 | 86.4 | 90.8 | 74.6 | 88.2 | 91.6 |
| NLPR3 | - | - | 95.6 | 93.0 | 85.6 | 90.3 | 84.5 | 89.8 | 91.2 |
| EaNet (Zhang et al., 2020) | CIR | $> 6 \cdot 10^7$ | 96.2 | 93.4 | 85.6 | 90.5 | 88.3 | 90.8 | 91.2 |
| CASIA2 (Liu et al., 2018) | CIR | $> 6 \cdot 10^7$ | 96.0 | 93.2 | 84.7 | 89.9 | 86.7 | 90.1 | 91.1 |
| DDCM (Liu et al., 2020) | CIR | $> 2 \cdot 10^7$ | 95.3 | 92.7 | 83.3 | 89.9 | 82.4 | 88.7 | 90.4 |
| DLR_9 (Marmanis et al., 2018) | CIR, nDSM | $8.1 \cdot 10^8$ | 95.2 | 92.4 | 83.9 | 89.9 | 81.2 | 88.5 | 90.3 |
| VFuseNet (Audebert et al., 2018) | CIR, nDSM | $> 8 \cdot 10^7$ | 94.4 | 91.0 | 84.5 | 89.9 | 86.3 | 89.2 | 90.0 |
| *FuseNet-lite* | CIR, nDSM | $4.6 \cdot 10^5$ | 95.6 | 92.4 | 85.0 | 90.1 | 83.5 | 89.3 | 90.9 |
| **Potsdam** | | | | | | | | | |
| HUSTW (Sun et al., 2019) | RGB, IR, nDSM | $1.3 \cdot 10^7$ | 96.7 | 93.8 | 88.0 | 89.0 | 96.0 | 92.7 | 91.6 |
| BAMTL (Wang et al., 2021) | RGB, nDSM | $> 6 \cdot 10^7$ | - | - | - | - | - | 90.9* | 91.3 |
| CASIA2 (Liu et al., 2018) | CIR | $> 6 \cdot 10^7$ | 97.0 | 93.3 | 87.7 | 88.4 | 96.2 | 92.5 | 91.1 |
| DDCM (Liu et al., 2020) | RGB | $> 2 \cdot 10^7$ | 96.8 | 93.3 | 87.6 | 89.4 | 95.0 | 92.4 | 91.1 |
| VFuseNet (Audebert et al., 2018) | RGB, IR, nDSM | $> 8 \cdot 10^7$ | 96.3 | 92.7 | 87.3 | 88.5 | 95.4 | 92.0 | 90.6 |
| *FuseNet-lite* | RGB, IR, nDSM | $4.6 \cdot 10^5$ | 97.1 | 92.7 | 87.5 | 87.8 | 94.9 | 92.0 | 90.6 |

Table 4: Comparison to state-of-art with eroded reference. Mean F1 score with * considers class *clutter* as well. #Param.: number of trainable parameters. For the method described by NLPR3 in the scoreboard, no publication is found. "-": no value is given.

In both test sites, the best OA of 91.6% are delivered by the method described as HUSTW. Although *FuseNet-lite* does not outperform this method, the difference in OA achieved by *FuseNet-lite* is within 1% of the best results. In Vaihingen, the predictions are quite close to the best ones in terms of both OA and mF1 score. *FuseNet-lite* outperforms HUSTW in terms of the mean F1 score by about 1%, mainly due to the improved recognition of class *car*. In Potsdam, there is no large difference (-0.7%) between our mean F1 score and the best one, yet a difference of 1% in OA is observed. Turning the focus on the number of trainable parameters, our network is much slimmer than all others by a factor of about 100. All works mentioned in the table are based on encoder-decoder network structures and most of them use an encoder based on ResNet (He et al., 2016). For instance, EaNet, CASIA2 and BAMTL are based on ResNet101; DDCM is based on ResNet50 and VFuseNet is based on ResNet34. The decoder part is symmetric to the encoder. Due to the large number of trainable parameters in ResNet, they suffer from having to determine a large number of unknown parameters from the training data. Generally speaking, *FuseNet-lite* requires only about 1% or even fewer trainable parameters than the compared methods. In conclusion, our method delivers predictions on par with the state-of-art methods, yet requires a considerably smaller number of trainable parameters.

The results discussed so far are based on an evaluation using the eroded reference to allow for a comparison to other methods. When using the full reference for evaluation, the OA and mean F1 scores (without considering the *clutter* class) are 87.9% and 85.9% in Vaihingen and 88.5% and 89.5% in Potsdam, respectively. In terms of OA these numerical values are close to those in Hameln and Schleswig, indicating a consistently good performance of our method.

**5.2. Evaluation of land use classification**

**5.2.1. Comparison of network variants on the basis of individual training**

Tab. 5 presents an overview of the results of land use classification of all network variants in the two test sites (OA and mean F1)[1]. The table reveals that the variants based on joint optimization outperform the variants *LuNet-lite-MT* and *LuNet-lite-F2C* in most cases, delivering the best results in terms of OA and mean F1 score in both test sites in most cases. The variant *LuNet-lite-MT* delivers better classification results than *LuNet-lite-F2C* in terms of OA and mean F1 score over all semantic levels, but the results can be contradictory to the hierarchical class catalogue. In the classification results of Hameln, we found that about 7.3% predictions are non-consistent with the hierarchy, whereas there are about 9.5% such predictions in Schleswig. *LuNet-lite-F2C* achieves consistency with the class catalogue, but at the price of a somewhat lower performance, especially in level II. It may come as a surprise that there is a difference in the predictions of variants *LuNet-lite-MT* and *LuNet-lite-F2C* at level III, because *LuNet-lite-F2C* uses the raw output of *LuNet-lite-MT* for class prediction at that level. These differences can be attributed to the use of slightly different training strategies (Yang et al., 2020b).

Both variants based on joint optimization achieve a better accuracy than *LuNet-lite-MT* (up to +1.1% in OAs and up to + 2.9% in mF1) over level II and level III while at the same time delivering a consistent result. At level I, their differences are quite small. Comparing the variants based on joint optimization to *LuNet-lite-F2C*, the improvement of OA ranges from 0.4% (level I) to 1.6% (level II and level III) in Hameln. Besides, the discrimination of some individual classes is improved somewhat more, the largest improvement of mean F1 score being 2.7% at level III in Hameln. A similar picture of improvement is also seen in Schleswig, where the OA is improved between 1.1% (level II) and 1.6% (level III) for the joint optimization, and the mean F1 score from 1.6% (level I) to 4.3% (level II). In both sites, the improvement of the mean F1 score is larger than the one of OA. As pointed out in section 3.2.5, the main advantage of this strategy is that it can mitigate the problems that occur when the probabilities of two classes are too close, thus, leading to wrong predictions in F2C. Thus, the classification results at multiple levels support each other, while in the greedy scheme employed in *LuNet-lite-F2C* a poor classification in level III will affect the other levels in a negative way.

---

[1] There is no comparison to *LuNet-MT* and *LuNet-F2C* separately, because the decrease of the number of parameters is assumed to have a very similar effect to the one of *LuNet-JO* vs. *LuNet-lite-JO*.

| Test site | Network variant | Category level | | | | | |
|---|---|---|---|---|---|---|---|
| | | level I | | level II | | level III | |
| | | OA [%] | mF1 [%] | OA [%] | mF1 [%] | OA [%] | mF1 [%] |
| Hameln | *LuNet-lite-MT* | **92.7** | **86.7** | 78.2 | 61.1 | 73.8 | 51.2 |
| | *LuNet-lite-F2C* | 92.1 | 86.3 | 77.1 | 59.8 | 72.5 | 51.4 |
| | *LuNet-lite-JO* | 91.5 | 85.9 | 78.5 | 59.9 | **74.1** | 51.8 |
| | *LuNet-JO* | 92.5 | 86.5 | **78.7** | **61.8** | 74.1 | **54.1** |
| Schleswig | *LuNet-lite-MT* | 90.8 | 85.7 | 73.7 | **63.9** | 69.8 | 57.2 |
| | *LuNet-lite-F2C* | 90.0 | 84.2 | 73.7 | 59.6 | 69.3 | 53.8 |
| | *LuNet-lite-JO* | 91.0 | 85.4 | **74.8** | 61.8 | 70.4 | 55.3 |
| | *LuNet-JO* | **91.4** | **85.8** | **74.8** | **63.9** | **70.9** | **58.0** |

Table 5: Overview of the results of hierarchical land use classification for the four network variants discussed in this paper. The networks were trained using samples from the corresponding dataset only. mF1: mean F1 score; OA: overall accuracy. Best scores per test site are shown in bold font. Results are given independently for the three semantic levels I to III.

Comparing the results achieved by all variants, it is obvious that the classification accuracy decreases with increasing the semantic resolution, as could be expected. At the coarsest level (I), the OA is about 90% for all variants. It would seem that at this level, the CNN-based classification outperforms the CRF-based method of Albert et al. (2017) (OA 78%), but a direct comparison is not possible because the class structures are not identical. At the intermediate level (II), the OA drops by 15%-20% compared to level I. The fact that the drop in the mean F1 scores is even larger indicates that a non-negligible number of classes can no longer be differentiated. At the finest level, the performance is even worse, with an additional decrease of about 5% in OA compared to level II. Again, the drop in the mean F1 scores is larger (about 8%). There are two main reasons for these problems at the finer levels. On the one hand, the number of training samples of some classes is very low, which may lead to an insufficient representation of these categories (cf. Tab. 1). Second, there are quite a few object types derived from the same ancestor category which are quite similar in appearance, shape and composition of land cover types, so that it becomes difficult to differentiate between them even for a human. For instance, class *industry* in level II is very similar to *residential* with dense buildings and sealed streets.

Finally, we compare the results of the two network variants based on joint optimization, i.e. *LuNet-JO* and *LuNet-lite-JO*. The results are relatively similar, with *LuNet-JO* performing better in level I and level II (up to +1% in OA and +1.9% in mean F1 score); in level III *LuNet-JO* achieves better results in mean F1 score (+2.3%). In Schleswig, *LuNet-JO* performs also better than *LuNet-lite-*

*JO* in nearly all indices: up to +0.5% in OA and +2.7% in mean F1 score over all levels.

Fig. 10 shows the training and validation errors as functions of the training epoch for one group from 6-fold cross validation in Schleswig. The error gap (*training error - validation error*) differs according to the hierarchy level. In level I, both networks can achieve a small error gap (about 4% in *LuNet-JO* and 3% in *LuNet-lite-JO*). As the number of categories increases, the error gap becomes larger in both networks. On average, the error gap is decreased by about 2.5% in level II and 3% in level III by using *LuNet-lite-JO* compared to *LuNet-JO*. We take these findings as an indication that reducing the number of unknown parameters mitigates overfitting (as indicated by the error gap) to some degree. This observation and the fact that *LuNet-lite-JO* delivers results that are on par with those achieved by *LuNet-JO* leads us to preferring the former over the latter. Thus, in the subsequent sections, we use *LuNet-lite-JO* for further investigations.

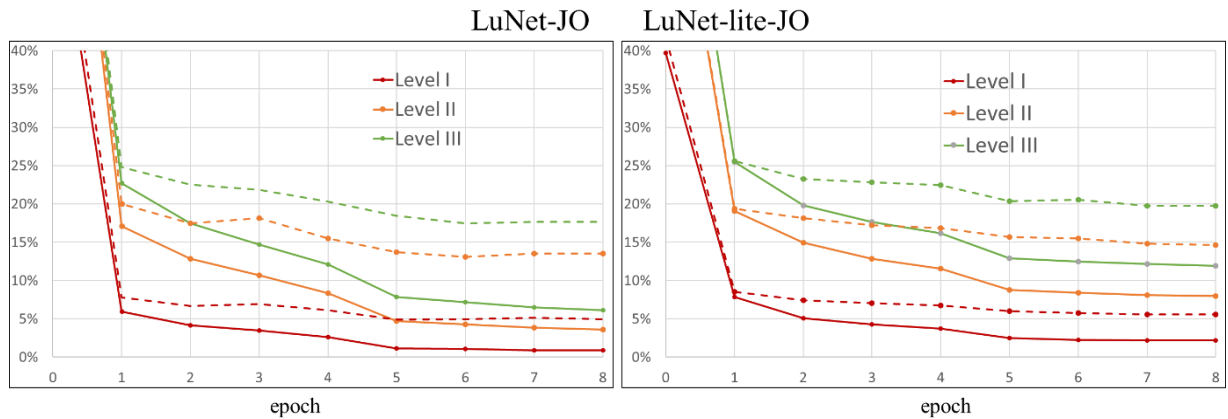

Figure 10: Training and validation errors (100% - OA) of hierarchical land use classification in Schleswig as a function of the training epoch number (group 1 in 6-fold cross validation). Solid lines: training error; dashed lines: validation error. The corresponding curves for the other groups show a similar behaviour and are omitted for lack of space.

**5.2.2. Comparison of training strategies: individual vs. combined training**

In section 5.2.1, all methods are trained and tested individually for the two test datasets. Here, we combine the data of both test sites for further testing. We know that there is a domain gap between the datasets, mainly caused by different capturing seasons (spring in Hameln, summer in Schleswig), which should mainly affect vegetation objects, e.g. deciduous trees and agriculture. Nevertheless, combining the datasets is expected to help in the identification of some categories, because it increases the number of training samples. This would be particularly relevant for the under-represented categories at the finest level, e.g. *extended residential*, *mixed usage*, *hardwood and softwood, stagnant waterbodies* in Hameln, or *factory, flowing waterbodies* in Schleswig.

An overview of the results when applied to the combined dataset is shown in Tab. 6, along with a comparison to those achieved by individual training and testing. In Tab. 7 detailed F1 scores of individual classes from individual and combined training are presented. We start with an analysis of the results achieved for test samples from Hameln before switching our attention to Schleswig.

| Test site | Training | Category level | | | | | |
|---|---|---|---|---|---|---|---|
| | | level I | | level II | | level III | |
| | | OA [%] | mF1 [%] | OA [%] | mF1 [%] | OA [%] | mF1 [%] |
| Hameln | *Individual* | 91.5 | 85.9 | **78.5** | 59.9 | 74.1 | 51.8 |
| | *Combined* | **92.8** | **88.5** | 78.4 | **62.3** | **74.5** | **54.5** |
| Schleswig | *Individual* | 91.0 | 85.4 | 74.8 | 61.8 | 70.4 | 55.3 |
| | *Combined* | **91.8** | **86.7** | **75.2** | **62.1** | **70.7** | **56.7** |

Table 6: Results of hierarchical land use classification for *LuNet-lite-JO* when trained using only one test site (*Individual* training) and a combined dataset consisting of samples from both test sites. mF1: mean F1 score; OA: overall accuracy. The results for individual training are identical to those achieved for *LuNet-lite-JO*, see Tab. 5; they are included here for a better comparison. Best scores per test site are shown in bold font.

| level I | | | | | level II | | | | | level III | | | | |
|---|---|---|---|---|---|---|---|---|---|---|---|---|---|---|---|
| Category | Hameln | | Schleswig | | Category | Hameln | | Schleswig | | Category | Hameln | | Schleswig | |
| | IND | COM | IND | COM | | IND | COM | IND | COM | | IND | COM | IND | COM |
| settlement | 92.1 | **93.5** | 93.3 | 93.9 | res. | 86.9 | 87.3 | 87.1 | 86.6 | res.use | 89.7 | 89.0 | **88.4** | 87.4 |
| | | | | | | | | | | ext. res. | 36.1 | **57.0** | 68.4 | **69.4** |
| | | | | | ind. | 69.7 | **72.6** | 57.9 | **58.8** | fact. | **34.4** | 29.4 | 0 | **14.3** |
| | | | | | | | | | | busi. | 52.2 | 52.8 | **53.9** | 52.0 |
| | | | | | | | | | | infra. | 29.9 | **49.0** | 32.4 | **45.5** |
| | | | | | mix | 0 | 0 | 27.6 | **28.8** | mix | 0 | 0 | 27.6 | **28.8** |
| | | | | | special | 52.2 | 52.6 | **38.6** | 37.3 | special | 52.2 | 52.6 | **38.6** | 37.3 |
| | | | | | rec. | 75.5 | **78.1** | 68.9 | **71.4** | leis. | **6.2** | 0 | **52.3** | 49.9 |
| | | | | | | | | | | park | 76.7 | **78.9** | 68.6 | 68.8 |
| traffic | 92.5 | **93.6** | 90.5 | **91.7** | ro.traf. | **83.6** | 82.6 | 84.8 | 84.7 | mo. road | **86.4** | 85.1 | 88.7 | 88.4 |
| | | | | | | | | | | traf. area | 63.7 | **69.1** | 42.3 | **44.4** |
| | | | | | path | **84.7** | 82.8 | 72.5 | **73.5** | path | **84.7** | 82.8 | 72.5 | **73.5** |
| | | | | | park.lot | 50.9 | **58.0** | 26.9 | **32.3** | park.lot | 50.9 | **58.0** | 26.9 | **32.3** |
| vegetation | 83.0 | **85.8** | 93.6 | 94.0 | agr. | 87.4 | 86.6 | 92.0 | **93.3** | farm | 85.4 | **90.5** | 90.1 | **95.6** |
| | | | | | | | | | | garden | 68.7 | **71.2** | 85.9 | **88.0** |
| | | | | | forest | **84.0** | 81.3 | 89.5 | 89.1 | h/s.wood | 59.9 | 58.5 | 46.0 | 45.8 |
| | | | | | | | | | | h&s.wood | **47.6** | 29.5 | **58.3** | 56.1 |
| | | | | | grove | **54.0** | 43.0 | **58.6** | 49.0 | grove | **54.0** | 43.0 | **58.6** | 49.0 |
| | | | | | moor | 26.7 | 26.7 | 63.1 | **64.9** | moor | 26.7 | 26.7 | 63.1 | **64.9** |
| water | 76.1 | **81.1** | 64.2 | **67.2** | flow.wat. | 74.1 | **75.9** | 26.0 | **31.3** | flow.wat.bo | 74.1 | **75.9** | 26.0 | **31.3** |
| | | | | | stag.wat. | 9.2 | **44.9** | 72.4 | 68.4 | stag.wat.bo. | 9.2 | **44.9** | 72.4 | 68.4 |

Table 7: F1 scores of individual categories of all levels from *LuNet-lite-JO*. IND: results from individual training; COM: results from combined training. In each test site, for one specific category, if the absolute difference of the F1 scores between IND and COM training are more than 1%, the best score is shown in bold font.

In Hameln, the OAs of the two training variants are close to each other for all three semantic levels, with an absolute difference smaller than 1.3%. Nonetheless, the mean F1 score of all levels achieved by combined training is better than those achieved by individual training by at least 2.4% (level II). Looking at the individual F1 scores in Tab. 7, it is clear that most categories derived from *settlement* and *traffic* are differentiated better in the combined training setup. The reason could be that the difference in the appearance of the objects of these categories in the two datasets is relatively small, thus, more data helps their identification. However, compared to individual training, there are drops up to 7.5% in terms of the F1 score for the categories *factory*, *business* and *residential* when more data of these two categories are added for training. The fact that the F1 scores of these classes are relatively low may be attributed to some similarity between objects of these categories between each other. For instance, according to the confusion matrix, about 15% of the samples belonging *business* are classified as *residential* and 10% as *factory*; similarly, about 51% *factory* samples are classified as *business*. It remains unclear why combining the training samples from the two test sites reduces the corresponding F1 scores.

The categories *farm land* and *garden land*, both derived from *vegetation* are correctly identified much better in the combined training setup where the largest increase of F1 score is 5.1% for *farm land*. It would seem that despite the seasonal differences to be expected for these objects, the availability of a large number of samples in the combined dataset supports the differentiation of these classes in Hameln. However, for categories *hardwood and softwood* and *grove*, which mainly consist of deciduous trees, the separability is decreased in the combined setup, with a drop of F1 score of up to 18.1%.

Switching the focus on Schleswig, the combined training delivers almost the same performance as the individual training in terms of OA over all semantic levels. Similar to Hameln, the largest increase can be observed in the mean F1 score of level III (+1.4%). The individual F1 scores per class show that more than half the classes in level III are classified equally well or better after combining the datasets. The largest increase of F1 score is achieved for the category *factory* (+14.3%), whereas it cannot be correctly identified in the individual training setup. Therefore, it can be concluded that adding more different data helps the classification, especially for the under-represented categories.

### 5.2.3. Detailed analysis of *LuNet-lite-JO* in combined training

In this section, we analyse the results achieved by *LuNet-lite-JO* in the combined training setup level by level to obtain further insights into the separability of classes as a function of the semantic level of detail. The analysis is carried out on the basis of the results shown in Tab. 7.

**Level I**: in this level, the four classes can be separated easily in both datasets. However, in both cases, mean F1 scores of less than 82% for the class *water body*, indicate a problem with that class. This may partly be due to the fact that there are very few samples of that class (about 2.8% in the combined training dataset). Furthermore, the confusion matrix shows that about 18% of the samples of *water body* in Hameln are confused with *traffic*; the corresponding percentage for Schleswig is 22%. The reason for this confusion could be that both object types are very similar in shape and some land cover components (e.g. both are surrounded by *grass* and *tree* pixels), and the main material (*water* vs. *sealed*) appears similar, may be occluded or affected by cast shadow. In combination with the relatively small number of training samples for water, these problems apparently prevent the CNN from properly learning to differentiate these classes.

**Level II:** we analyse the results at this level according to the ancestor categories in level I. There are only two level II sub-categories of **settlement** achieving F1 scores over 60% in both data sets (*residential* and *recreation*); obviously, it is more difficult to differentiate the other categories correctly. The main source of errors is a confusion between *mixed usage* and *industry*. Again, this may be due to their similar appearance and composition of land cover elements (cf. Fig. 11 a and 11 b). Among the sub-categories of **traffic**, *road traffic* and *path* are differentiated more easily (F1 scores ≥ 73% in both sites). *Parking lot* is frequently confused with *road traffic* and *industry*. It can be differentiated much better in Hameln than in Schleswig, where about 44% of the *parking lot* objects are erroneously assigned to *road traffic*. Figs. 11 b and 11 c show examples for two objects belonging to the classes *industry* and *parking lot* that have a very similar appearance; this degree of similarity between these object types may be the reason why they are confused frequently. Among the sub-categories of **vegetation**, *agriculture* and *forest* are particularly well classified (F1 ≥ 81%) in both cases. The other sub-categories are more problematic. *Grove* is most frequently confused with *recreation* and *forest*, i.e. again with classes that have a similar appearance (cf. Figs. 12 a and 12 b). The category *moor* is mainly confused with *agriculture*.

**Level III**: in level III, some classes can be differentiated very well, e.g. *residential in use* or *motor road*, both with F1 scores larger than 85% in all cases. Although more than half of the categories achieve F1 scores larger than 50%, it is still more difficult to separate them than those of the other levels. We think that to a large degree this is due to the comparably small number of training samples for some classes, even after merging the datasets.

In summary, with an increasing number of categories from level to level, it becomes more and more difficult to differentiate them. At the finer levels, some categories are very similar in appearance and land cover composition (e.g. *industry* vs. *mixed usage*; *grove* vs. *forest*), which may make them difficult to be discerned. On the other hand, we believe that the scarcity of training samples has a negative impact on some classes. Thus, although this will not solve all remaining problems, we think that the situation could be improved by increasing the number of training samples further, e.g. by enlarging the size of the area to be processed. This would certainly be required if the classification at level III should achieve the level of accuracy required for operational use.

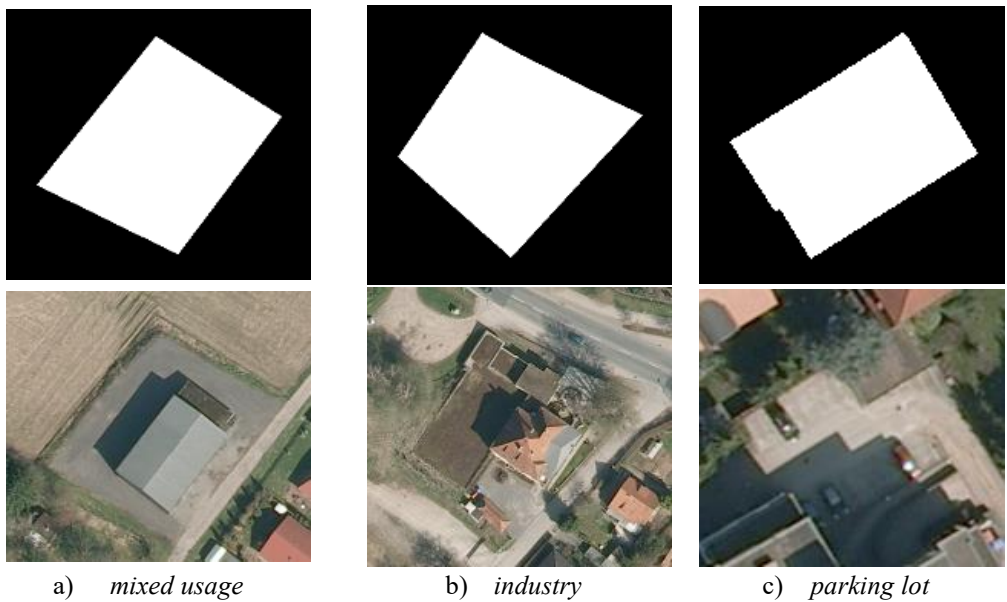

a)  *mixed usage*          b)  *industry*          c)  *parking lot*

Figure 11: Similar land use objects in category level II corresponding to *settlement* in level I. Upper row: binary images indicating the object shapes, bottom row: RGB orthophoto. The images are rescaled for visualization.

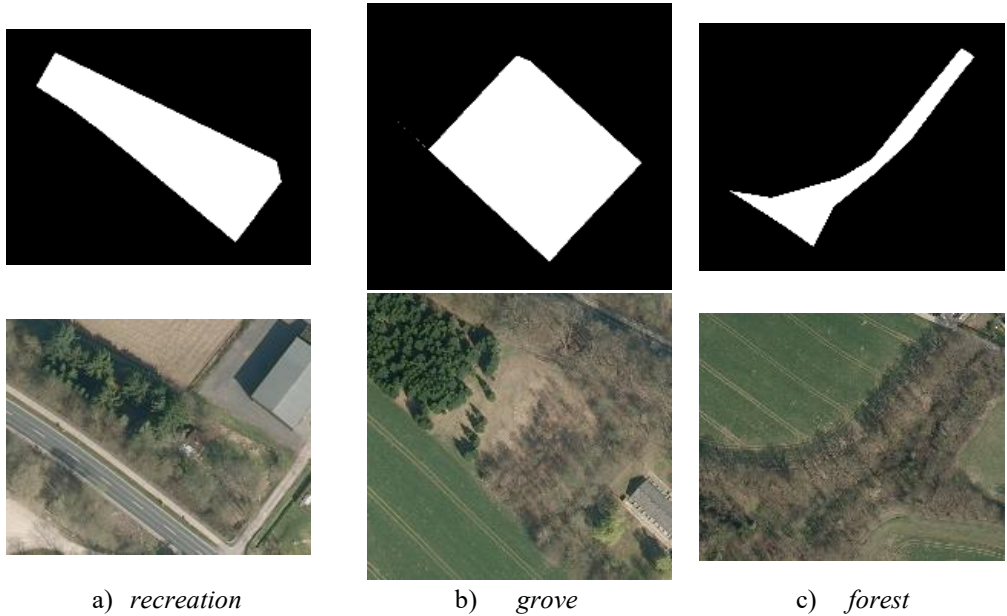

a) *recreation*  b) *grove*  c) *forest*

Figure 12: Similar land use objects in category level II corresponding to *settlement* and *vegetation* in level I. Upper row: binary images indicating the object shapes, bottom row: RGB orthophoto. The images are rescaled for visualization.

### 5.2.4. Influence of the object size

Having noted a large influence of the object size on the classification accuracy in our previous work (Yang et al., 2018; 2019), we also analyse this topic for the new hierarchical land use classification scheme based on joint optimization. For that purpose, we use the results from the combined training of network variant *LuNet-lite-JO*. Larger objects, i.e. objects that have to be split into tiles (cf. section 3.2.2), would be expected to be better identified, because the predicted combined class label can be interpreted as the result of an ensemble of multiple predictions (one per tile; cf. section 3.2.5). Fig. 13 shows the overall accuracy and mean F1 score as a function of object size. As far as the OA is concerned, in both datasets, the expected trend is clearly visible for all levels, despite the use of the ROI layer that was supposed to mitigate the problems with small objects (Yang et al., 2019). In Hameln, even small objects can be detected with an OA better than 90% at level I, but the larger the object, the better the prediction becomes. The differences of OA between objects of different size are more pronounced in levels II and III. The most significant improvement of OA occurs between objects having a size between $A = 2621$ m$^2$ (the area of a patch of 256 x 256 pixels at a GSD of 20 cm) to 2A and objects of size 2A to 3A (+ 9% in level II and + 7% in level III). In Schleswig, there is a similar tendency. It seems to be more difficult to classify objects smaller than size A in level I, and in the other levels, the biggest increase occurs between objects smaller than A and objects having an area between A to 2A (+7% in level I, +14% in level II and +13% in level III). Considering the mean F1 scores in both sites, the tendency to increase with the area can be observed at level I, similar to the case of OA. However, at level II the mean F1 scores does not change too much with object size in both sites, which is mainly due to the unbalanced class distribution, which affects both large and small objects. At level III, the mean F1 scores for objects of different size are very similar in Hameln. In Schleswig, it increases continuously until a size of 3A. For objects having size larger than 3A it is slightly lower than the one of objects of 2A to 3A, but the difference is small (about 2.5%).

This analysis makes clear that in addition to the number of training samples per class, the object size is another limiting factor for the success of the classification at all semantic levels. Whereas at level I, the accuracies achieved for small objects (~85%-90%) may still be in an acceptable range, at higher semantic levels this problem becomes more crucial, e.g. leading to about 40% of false classifications for small objects in Schleswig.

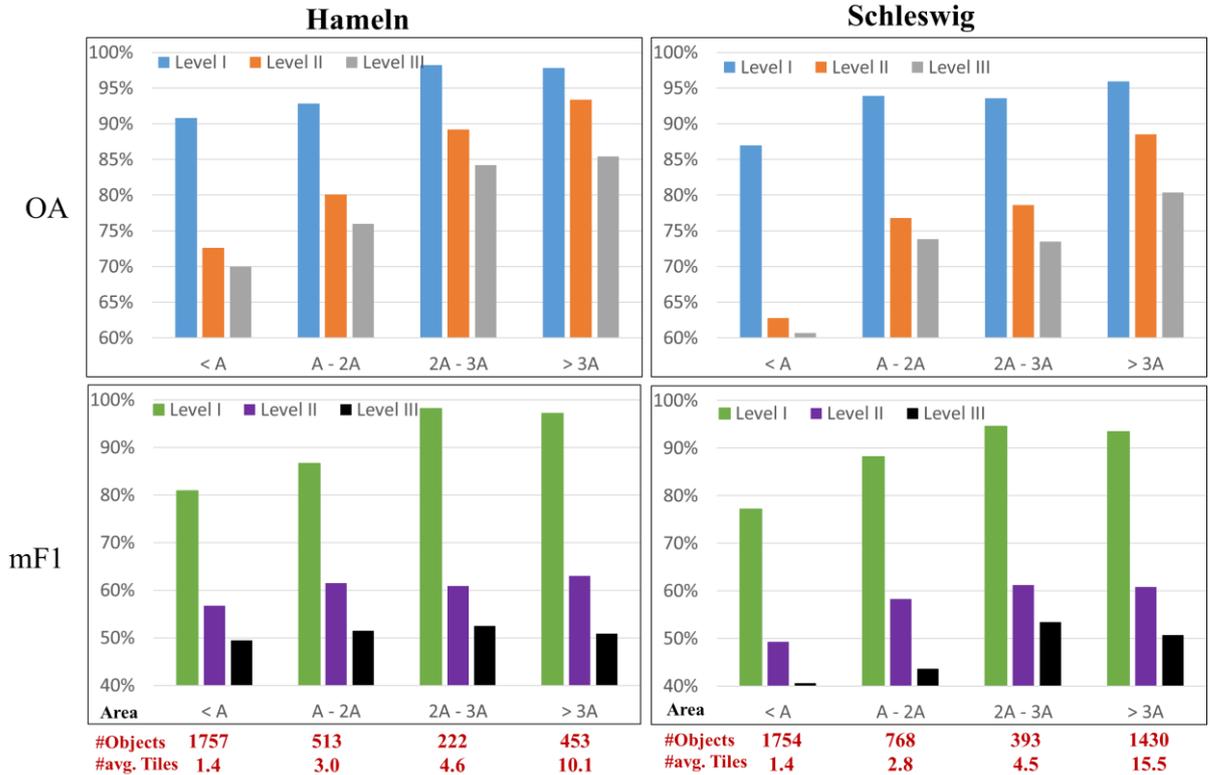

Figure 13: Overall accuracy (upper row) and mean F1 score (bottom row) of land use classification achieved by *LuNet-lite-JO* with combined training as a function of object size given in units of A = 2621 m$^2$, i.e. the area of a patch of 256 x 256 pixels at a GSD of 20 cm. The number of objects and the corresponding average number of tiles in each area bin are shown in red at the bottom of the figure. Note that an object that fits into a tile according to its area might still have to be tiled, e.g. if it has an elongated shape. The scaling of the Y axis was chosen for better interpretability.

## 6. Conclusion

In this paper, we have proposed a hierarchical deep learning framework for predicting land use labels of objects a geospatial database from high resolution images with the final goal of verifying the database as a first step of an update process. For this purpose, a two-step strategy is applied in which we firstly obtain pixel-level land cover information for the given input data, and then use the results together with the images to classify land use objects. The main findings can be summarized as follows:

- The new joint optimization and training strategy proposed to achieve consistent hierarchical land use classification with respect to the object class hierarchy outperforms our baseline methods, indicating that a joint consideration of class scores over all semantic levels improves the classification results. At the same time, and unlike the baseline methods, it guarantees consistency with the hierarchical object class catalogue. In our previous work (Yang et al., 2020b), this consistency was enforced at the cost of classification accuracy at the coarser level. In contrast, for the new strategy the classification tasks at different semantic levels were found to stabilise each other.

- Using "light" variants of our CNNs for land cover and land use classification, reducing the number of unknown parameters by more than 90%, we found that the classification accuracy was hardly affected, but the degree of overfitting was reduced.

- We combine datasets to classify land cover and land use to see whether using the combined set for training could deliver better results despite a considerable domain gap between the individual constituents. Both for land cover and land use classification, combining the datasets has a positive impact on the classification, which is in line with the findings in (Kaiser et al., 2017). We believe this to be due to the increased the number of available samples for these classes.

- Comparing the classification accuracy achieved at different semantic levels, the categories at the coarsest level can be discerned best, with overall accuracies in the order of 90%. As the semantic resolution and, accordingly, the number of categories is increased, the classes are harder to be differentiated correctly, with a drop in overall accuracy in the order of 15% for level II and a further 5% for level III. The main reasons seem to be that at the finer levels, the number of training samples per class is much smaller and there are more categories that are quite similar in appearance.

- Object size could be confirmed as an additional factor influencing the classification accuracy: it is more difficult to classify small objects correctly than larger ones.

Future work on land cover classification will concentrate on assessing the impact of enlarging the training set to be able to differentiate a more diverse land cover class structure. In order to reduce the efforts for manual labelling when enlarging the training set, existing land cover databases should be applied to derive the class labels. Such a procedure requires a strategy to mitigate the impact of wrong class labels of training samples (label noise) that occur due to temporal changes and other reasons. One way to accomplish this task is to transfer the principles for dealing with label noise developed in (Maas et al., 2019) to CNN-based classification.

Future work on land use classification will try to improve the classification accuracy of small objects, for instance by increasing the level of data augmentation for small objects or adapting the loss functions such that small objects obtain larger weights and, thus, impact on the learned CNN. Considering multi-scale approaches may also mitigate this problem. Furthermore, we currently consider the training labels for land use objects to be correct, which can only be achieved by checking and, if required, adapting the class labels of the objects in an interactive and, thus, time-consuming process. As we identified the number of training samples to be a major limiting factor of classification accuracy at finer semantic levels, it would be desirable to expand the training dataset. This could be achieved by using the entire geospatial database for training, using the class information from the reference to define the training labels. Again, due to temporal changes or errors in the original acquisition, a certain proportion of these labels are be wrong, i.e. also affected by label noise. Thus, label noise robust training strategies tailored for this task also need to be developed. Finally, it would be interesting to combine land cover and land use classification in a unified CNN-based approach, leading to simultaneous classification of both tasks using an end-to end-trained network, e.g. (Zhang et al., 2019).

Finally, the current strategy for database verification relies on the object boundaries stored in that database. Errors in these boundaries could only be detected if they are large enough to lead to contradictions between the predicted land use of an object and the value stored in the database. Other directions of future research focus on an improved detection of object boundaries and of changes in those boundaries. This might be achievable by training a CNN to predict object boundaries, extending the works of (Marmanis et al., 2018) and (Volpi & Tuia, 2018) to the boundaries of land use objects.

**Acknowledgement**

We thank the Landesamt für Geoinformation und Landesvermessung Niedersachsen (LGLN), the Landesamt für Vermessung und Geoinformation Schleswig Holstein (LVermGeo) and the Landesamt für innere Verwaltung Mecklenburg-Vorpommern (LaiV-MV) for providing the test data and for their support of this project.